\documentclass{article}

\usepackage{iclr2025_conference,times}

\usepackage[textsize=tiny]{todonotes}

\newcommand{\zkn}[1]{}
\newcommand{\ky}[1]{}
\newcommand{\gc}[1]{}
\newcommand{\asn}[1]{}
\newcommand{\aen}[1]{}
\newcommand{\asz}[1]{}

\usepackage[utf8]{inputenc} %
\usepackage[T1]{fontenc}    %
\usepackage{url}            %
\usepackage{booktabs}       %
\usepackage{amsfonts}       %
\usepackage{nicefrac}       %
\usepackage{microtype}      %
\usepackage{xcolor}         %
\usepackage{amsmath}
\usepackage{amssymb}
\usepackage{mathtools}
\usepackage{amsthm}
\usepackage{enumitem}
\usepackage{xspace}
\usepackage{colortbl}
\usepackage{bbm}
\usepackage{subcaption}
\usepackage{wrapfig}

\definecolor{citecolor}{HTML}{2779af}
\definecolor{linkcolor}{HTML}{c0392b}
\usepackage[pagebackref=false,breaklinks=true,colorlinks,bookmarks=false,citecolor=citecolor,linkcolor=linkcolor]{hyperref}

\usepackage[capitalize]{cleveref}
\newcommand{\Method}{Reinforcement Learning In Context \xspace}
\newcommand{\method}{ReLIC\xspace}
\newcommand{\rls}{RL$^{2}$\xspace}
\newcommand{\semethod}{Transformer-SE\xspace}
\newcommand{\ieafmethod}{\method-No-IEA\xspace}
\newcommand{\trxl}{Transformer-XL\xspace}
\newcommand{\task}{\textsc{ExtObjNav}\xspace}
\newcommand{\obnav}{\textsc{ObjectNav}\xspace}
\newcommand{\Task}{Extended Object Navigation\xspace}

\newcommand{\sinkv}{Sink-KV\xspace}

\title{\method: A Recipe for 64k Steps of In-Context Reinforcement Learning for Embodied AI}

\author{%
    Ahmad Elawady \quad
    Gunjan Chhablani \quad
    Ram Ramrakhya \quad
    Karmesh Yadav \quad \\
    \textbf{Dhruv Batra}  \quad
    \textbf{Zsolt Kira} \quad
    \textbf{Andrew Szot} \quad \\
    Georgia Tech 
}

\iclrfinalcopy %
\begin{document}

\maketitle

\begin{abstract}
\looseness=-1 Intelligent embodied agents need to quickly adapt to new scenarios by integrating long histories of experience into decision-making. For instance, a robot in an unfamiliar house initially wouldn't know the locations of objects needed for tasks and might perform inefficiently. However, as it gathers more experience, it should learn the layout of its environment and remember where objects are, allowing it to complete new tasks more efficiently. To enable such rapid adaptation to new tasks, we present \method, a new approach for in-context reinforcement learning (RL) for embodied agents. With \method, agents are capable of adapting to new environments using 64,000 steps of in-context experience with full attention while being trained through self-generated experience via RL. We achieve this by proposing a novel policy update scheme for on-policy RL called ``partial updates'' as well as a Sink-KV mechanism that enables effective utilization of a long observation history for embodied agents. Our method outperforms a variety of meta-RL baselines in adapting to unseen houses in an embodied multi-object navigation task. In addition, we find that \method is capable of few-shot imitation learning despite never being trained with expert demonstrations. We also provide a comprehensive analysis of \method, highlighting that the combination of large-scale RL training, the proposed partial updates scheme, and the Sink-KV are essential for effective in-context learning.
The code for \method and all our experiments is at \href{https://github.com/aielawady/relic}{github.com/aielawady/relic}.

\end{abstract}

\section{Introduction}

A desired capability of intelligent embodied agents is to rapidly adapt to new scenarios through experience.
An essential requirement for this capability is integrating a long history of experience into decision-making to enable an agent to accumulate knowledge about the new scenario that it is encountering.
For example, a robot placed in an unseen house initially has no knowledge of the home layout and where to find objects.
The robot should leverage its history of experiences of completing tasks in this new home to learn the home layout details, where to find objects, and how to act to complete tasks successfully.

To achieve adaptation of decision-making to new tasks, prior work has leveraged a technique called in-context reinforcement learning (RL) where an agent is trained with RL to utilize past experience in an environment~\citep{wang2016learning,team2023human,duan2016rl,grigsby2023amago,melo2022transformers}. 
By using sequence models over a history of interactions in an environment, these methods adapt to new scenarios by conditioning policy actions on this context of interaction history without updating the policy parameters.
While in-context RL has demonstrated the ability to scale to a context length of a few thousand 
agent steps~\citep{team2023human,grigsby2023amago}, this falls short of the needs of embodied AI where single tasks by themselves can span thousands of steps~\citep{szot2021habitat}. 
As a result, the agent cannot learn from multiple task examples because the context required for multiple tasks cannot be accommodated within the policy context.
Furthermore, prior work typically focuses on non-visual tasks~\citep{grigsby2023amago,melo2022transformers,ni2023transformers}, where larger histories are easier to incorporate due to the compact state representation.

In this work, we propose a new algorithm for in-context RL, which enables effectively utilizing and scaling to 64,000 steps of in-context experience in partially observable, visual navigation tasks. 
Our proposed method called \Method (\method), achieves this by leveraging a novel update and data collection technique for training with long training contexts in on-policy RL.
Using a long context for existing RL algorithms is prohibitively sample inefficient, as the agent must collect an entire long context of experience before updating the policy.
In addition, the agent struggles to utilize the experience from long context windows due to the challenge of learning long-horizon credit assignment and high-dimensional visual observations.
To address this problem, we introduce ``partial updates" where the policy is updated multiple times within a long context rollout over increasing context window lengths. 
We also introduce \sinkv to further increase context utilization by enabling more flexible attention over long sequences by adding learnable \textit{sink key and value} vectors to each attention layer.
These learned vectors are prepended to the input's keys and values in the attention operation. 
\sinkv stabilizes training by enabling the agent to not attend to low information observation sequences. 

\begin{figure*}[!t]
    \centering
    \includegraphics[width=\textwidth]{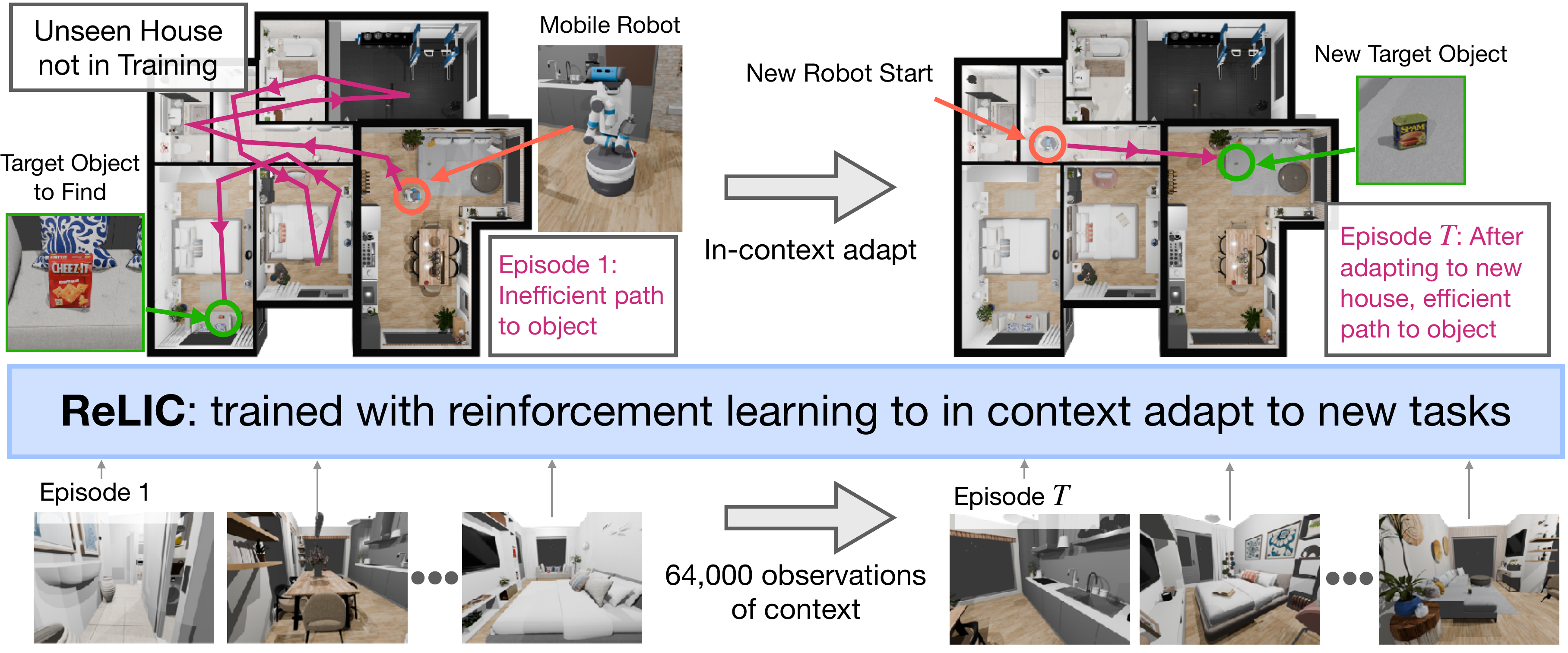}
    \caption{
        Overview of the \method approach and problem setup. \method learns a ``pixels-to-actions" policy from reward alone via reinforcement learning capable of in-context adapting to new tasks at test time. The figure shows the trained \method policy finding objects in an unseen house. In earlier episodes, the agent randomly explores to find the small target object since the scene is new. But after 64k steps of visual observations, \method efficiently navigates to new target objects.
    }
    \label{fig:teaser}
    \vspace{-10pt}
\end{figure*}

We test \method in a challenging indoor navigation task where an agent in an unseen house operating only from egocentric RGB perception must navigate to up to 80 \textit{small} objects in a row, which spans tens of thousands steps of interactions. 
\method is able to rapidly in-context learn to improve with subsequent experience, whereas state-of-the-art in-context RL baselines struggle to perform any in-context adaptation.
We empirically demonstrate that partial updates and \sinkv are necessary components of \method. 
We also show it is possible to train \method with 64k context length. 
Surprisingly, we show \method exhibits emergent few-shot imitation learning and can learn to complete new tasks from several expert demonstrations, despite only being trained with RL and never seeing expert demonstrations (which vary from self-generated experiences) during training.
We find that \method can use only a few demonstrations to outperform self-directed exploration alone. In summary, our contributions are:
\begin{enumerate}[itemsep=-0pt,topsep=0pt,parsep=0pt,partopsep=0pt,parsep=0pt,leftmargin=*]
    \item We propose \method for scaling in-context learning for online RL, which adds two novel components of partial updates and \sinkv. 
    We empirically demonstrate that this enables in-context adaptation of over 64k steps of experience in visual, partially observable embodied AI problems, whereas baselines do not improve with more experience.

    \item We demonstrate \method is capable of few-shot imitation learning despite only being trained with self-generated experience from RL.

    \item We empirically analyze which aspects of \method are important for in-context learning and find that sufficient RL training scale, partial updates, and the \sinkv modification are all critical.
\end{enumerate}

\section{Related Work}

\textbf{Meta RL.} Prior work has explored how agents can learn to quickly adapt to new scenarios through experience.
Meta-RL deals with how agents can learn via RL to quickly adapt to new scenarios such as new environment dynamics, layouts, or task specifications.
Since Meta-RL is a large space, we only focus on the most relevant Meta-RL variants and refer the readers to \citet{beck2023survey} for a complete survey of Meta-RL.
Some Meta-RL works explicitly condition the policy on a representation of the task and adapt by inferring this representation in the new setting~\citep{zhao2020meld,yu2020learning,rakelly2019efficient}.
Our work falls under the ``in-context RL'' Meta-RL paradigm where the policies implicitly infer the context by taking an entire history of interactions as input.
\rls~\cite{duan2016rl} trains an RNN that operates over a sequence of episodes with RL and the agent implicitly learns to adapt based on the RNN hidden state. 
Other works leverage transformers for this in-context adaptation~\citep{team2023human,melo2022transformers,laskin2022context}.
\cite{raparthy2023generalizationnewsequentialdecision,lee2023supervisedpretraininglearnincontext} also address in-context learning for decision making, but do so via supervised learning from expert demonstrations, whereas our work only requires reward.
Most similar to our work is AMAGO~\citep{grigsby2023amago}, an algorithm for in-context learning through off-policy RL.
AMAGO modifies a standard transformer with off-policy loss to make it better suited for long-context learning, with changes consisting of: a shared actor and critic network, using Leaky ReLU activations, and learning over multiple discount factors.
Our work does not require these modifications, instead leveraging standard transformer architectures, and proposes a novel update scheme and \sinkv for scaling the context length with on-policy RL.
Empirically, we demonstrate our method scaling to $8\times$ longer context length and on visual tasks, whereas AMAGO focuses primarily on state-based tasks.

\textbf{Scaling context length.}
Another related area of research scaling the context length of transformers.
Prior work extend the context length using a compressed representation of the old context, either as a recurrent memory or a specialized token \citep{dai2019transformerxl,munkhdalai2024leave,zhang2024soaring}.
Other work address the memory and computational inefficiencies of the attention method by approximating it \citep{beltagy2020longformer,wang2020linformer} or by doing system-level optimization \citep{dao2023flashattention2}. Another direction is context extrapolation at inference time either by changing the position encoding \citep{su2023roformer,press2022train} or by introducing attention sink \citep{xiao2023efficient}. Our work utilizes the system-level optimized attention \citep{dao2023flashattention2} and extends attention sinks for on-policy RL in Embodied AI.

\textbf{Embodied AI.} 
Prior work in Embodied AI has primarily concentrated on the single episode evaluation setting, where an agent is randomly initialized in the environment at the beginning of each episode and is tasked with taking the shortest exploratory path to a single goal specified in every episode~\citep{wijmans2019dd,yadav2022ovrl}. In contrast,~\cite{wani2020multion} introduced the multi-ON benchmark, which extends the complexity of the original task by requiring the agent to navigate to a series of goal objects in a specified order within a single episode. Here, the agent must utilize information acquired during its journey to previous goals to navigate more efficiently to subsequent locations. Go to anything (GOAT)~\citep{chang2023goat}, extended this to the multi-modal goal setting, providing a mix of image, language, or category goals as input. In comparison, we consider a multi-episodic setting where the agent is randomly instantiated in the environment after a successful or failed trial but has access to the prior episode history.

\vspace{-5pt}
\section{Method}
\vspace{-5pt}

We introduce \Method (\method) which enables agents to in-context adapt to new episodes without any re-training.
\method is built using a transformer policy architecture that operates over a long sequence of multi-episode observations and is trained with online RL. 
The novelty of \method is changing the base RL algorithm to more frequently update the policy with increasingly longer contexts within a policy rollout and adding \sinkv to give the model the ability to avoid attending to low-information context.
\Cref{sec:problem} provides the general problem setting of adapting to new episodes.
\Cref{sec:architecture} details the transformer policy architecture.
\Cref{sec:learning} describes the novel update scheme of \method.
Finally, \Cref{sec:impl-details} goes over implementation details.

\subsection{Problem Setting}
\label{sec:problem} 
We study the problem of adaptation to new scenarios in the formalism of meta-RL~\citep{beck2023survey}. 
We have a distribution of training POMDPs $ \mathcal{M}_i \sim p(\mathcal{M})$, where each $ \mathcal{M}_i $ is defined by tuple $ \left( \mathcal{S}_i, \mathcal{S}_i^{0}, \mathcal{O}_i, \mathcal{A}, \mathcal{T}, \gamma, \mathcal{R}_i \right) $ for observations $ \mathcal{O}_i$, states $ \mathcal{S}_i$ which are not revealed to the agent, starting state distribution $ \mathcal{S}_i^{0}$, action space $ \mathcal{A}$, transition function $ \mathcal{T}$, discount factor $ \gamma$, and reward $ \mathcal{R}_i$. In our setting, the states, observations, and reward vary per POMDP, while the action space, discount factor, and transition function is shared between all POMDPs. 

From a starting state $ s_0 \sim \mathcal{S}_i^{0} $, a policy $ \pi$, mapping observations to a distribution over actions, is rolled out for an \emph{episode} which is a sequence of interactions until a maximum number of timesteps, or a stopping criteria. 
We refer to a \emph{trial} as a sequence of episodes within a particular $ \mathcal{M}_i$. The objective is to learn a policy $ \pi $ that maximizes the expected return of an episode.
At test-time the agent is evaluated on a set of holdout POMDPs. 

\subsection{\method Policy Architecture}
\label{sec:architecture} 

Similar to prior work~\citep{grigsby2023amago,team2023human}, \method implements in-context RL via a transformer sequence model that operates over a history of interactions spanning multiple episodes.
At step $ t$ within a trial, \method predicts current action $ a_t$ based on the entire sequence of previous observations $ o_1, \dots , o_t$ which may span multiple episodes.
In the embodied AI settings we study, the observation $ o_t $ consists of an egocentric RGB observation from the robot's head camera along with proprioceptive information and a specification of the current goal.
Each of these observation components are encoded using a separate observation encoding network, and the embeddings are concatenated to form a single observation embedding $ e_t$.
A causal transformer network \citep{vaswani2023attention} $ h_\theta$ inputs the sequence of embeddings $ h_\theta(e_1, \dots, e_t)$. 
From the transformer output, a linear layer then predicts the actions.

The transformer model $ h_\theta$ thus bears the responsibility of in-context learning by leveraging associations between observations within a trial.
This burden especially poses a challenge in our setting of embodied AI since the transformer must attend over a history of thousands of egocentric visual observations. 
Subsequent visual observations are highly correlated, as the agent only takes one action between observations. 
Knowing which observations are relevant to attend to in deciding the current action is thus a challenging problem.
In this work, we build our architecture around full attention transformers using the same architecture as the LLaMA language model~\citep{touvron2023llamaoa}, but modify the number of layers and hidden dimension size to appropriately reduce the parameter count for our setting. 

We also introduce an architectural modification to the transformer called \textbf{\sinkv} to improve the transformer's ability to attend over a long history of visual experience from an embodied agent.
Building off the intuition that learning to attend over a long sequence of visual observations is challenging, we introduce additional flexibility into the core attention operation by prepending the key and value vectors with a per-layer learnable sequence of \emph{sink KV vectors}.
Specifically, for an input sequence $ X \in \mathbb{R}^{n \times d}$ of $ n$ inputs of embedding dimension $ d$, the attention operator projects $ X$ to keys, queries and values notated as $ K, Q, V$ respectively and all elements of $ \mathbb{R}^{n \times d}$ where we assume all hidden dimensions are $d$ for simplicity. 
The standard attention operation computes $ \text{softmax}\left( \frac{Q K^\top}{\sqrt{d}} \right) V $.
We modify calculating the attention scores by introducing learnable vectors $ K_s, V_s \in \mathbb{R}^{s \times d} $ where $ s$ is the specified number of ``sinks''. 
We then prepend $ K_s, V_s$ to the $K, V$ of the input sequence before calculating the attention. 
Note that the output of the attention operation is still $ n \times d$, as in the regular attention operation, as the query vector has no added component.
We repeat this process for each attention layer of the transformer, introducing a new $ K_s, V_s$ in each attention operation.
\sinkv only results in $n_{layers} \times s \times d$ more parameters, which is $0.046\%$ of the 4.5M parameter policy used in this work.

\sinkv gives the sequence model more flexibility on how to attend over the input. 
Prior works observe that due to the softmax in the attention, the model is forced to attend to at least one token from the input \citep{off_by_one_softmax,xiao2023efficient}.
\sinkv removes this requirement by adding learnable vectors to the key and value.
In sequences of embodied visual experiences, this is important as attention heads can avoid attending over any inputs when there is no new visual information in the current observations. 
This flexibility helps the agent operate over longer sequence lengths.

\subsection{\method Learning}
\label{sec:learning} 

\method is updated through online RL, namely PPO~\citep{schulman2017proximal}.
However, for the agent to be able to leverage a long context window for in-context RL, it must also be trained with this long context window.
PPO collects a batch of data for learning by ``rolling out'' the current policy for a sequence of $ T$ interactions in an environment. 
To operate on a long context window spanning an entire trial, the agent must collect a rollout of data that consists of this entire trial.
This is challenging because, in the embodied tasks we consider, we seek to train agents on trials lasting over 64k steps, which consists of at least 130 episodes.  
As typical with PPO, to speed up data collection and increase the update batch size we use multiple environment workers each running a simulation instance that the policy interacts with in parallel. 
With 32 environment workers, this corresponds to $ \approx 130k$ environment steps between every policy update.
PPO policies trained in common embodied AI tasks, such as \obnav, have only $128$ steps between updates and require $ \approx 50k$ updates to converge (for 32 environment workers, 128 steps per worker between updates and 200M environment steps required for convergence)~\citep{yadav2022ovrl}. Executing a similar number of updates would require \method to collect $ \approx 6 $ billion environment interactions. 
\zkn{Need to explain the distinction between update intervals. Would be nice to have a diagram showing entire trial, multiple episodes within it, then which are used for updates (and then show batch axis). \\ Andrew: I think a figure will properly address this, so no need to explain it in the writing.}

\method addresses this problem of sample inefficiency by introducing a \emph{partial update scheme} where the policy is updated multiple times throughout a rollout.
First, at the start of a rollout of length $ T$, all environment workers are reset to the start of a new episode.
Define the number of partial updates as $ K$. 
At step $ i \in  [0, T] $ in the rollout, the policy is operating with a context length of $ i-1$ previous observations to determine the action at step $ i$. 
Every $ T/K$ samples in the rollout, we update the policy.
Therefore, at update $ N$ within the rollout, the agent has collected $ NT/K$ of the $ T$ samples in the rollout. 
The agent is updated using a context window of size $ NT/K$, however, the PPO loss is only applied to the final $ T/K$ outputs.
The policy is changing every $ T/K$ samples in the rollout, so the policy forward pass must be recalculated for the entire $ NT/K$ window rather than caching the previous $ (N-1)T/K$ activations. 
In the last update in the rollout, after collecting the last $T/K$ steps, we update the policy with the loss applied on all steps in the rollout. We refer to this step as \textit{full update}.
At the start of a new rollout, the context window is cleared and the environment workers again reset to new episodes.

\subsection{Implementation Details}
\label{sec:impl-details} 
The transformer is modeled after the LLaMA transformer architecture \citep{touvron2023llamaoa} initialized from scratch. Our policy uses a pretrained visual encoder which is frozen during training. A MLP projects the output of the visual encoder into the transformer.
We only update the parameters of the transformer and projection layers while freezing the visual encoder since prior work shows this is an effective strategy for embodied AI \citep{khandelwal2022simple,vc2023}.
For faster policy data collection, we store the transformer KV cache between rollout steps. 
To fit long context during the training in limited size memory, we used low-precision rollout storage, gradient accumulation \citep{huang2019gpipe} and flash-attention \citep{dao2023flashattention2}.
\asz{What do you mean by ``low-precision rollout"? Clarify what precision you use like ``all models use bfloat16 parameters". And why the distinction with rollout? Is the weight precision different between rollouts and learning? \\Ahmad: I clarified that it's the rollout storage. FYI, I explain the model's precision in the appendix. Mainly, the model is trained with FP32, except for the visual encoder which is FP16. The rollout storage is FP16.}
After each policy update, we shuffle the older episodes in the each sequence and update the KV-cache. 
Shuffling the episode serves as regularization technique since the agent sees the same task for a long time. It also reflects the lack of assumptions about the order of episodes, an episode should provide the same information regardless of whether the agent experiences it at the beginning or at the end of the trial.

We use the VC-1 visual encoder and with the ViT-B size \citep{vc2023}. 
We found the starting VC-1 weights performed poorly at detecting small objects, which is needed for the embodied AI tasks we consider.
We therefore finetuned VC-1 on a small objects classification task.
All baselines use this finetuned version of VC-1.
We provide more details about this VC-1 finetuning in \Cref{sec:visual-encoder-finetuning} and details about all hyperparameters in \Cref{sec:hyperparams}.

\begin{figure*}[t]
  \centering
  \begin{subfigure}[t]{0.45\columnwidth}
    \includegraphics[width=\textwidth]{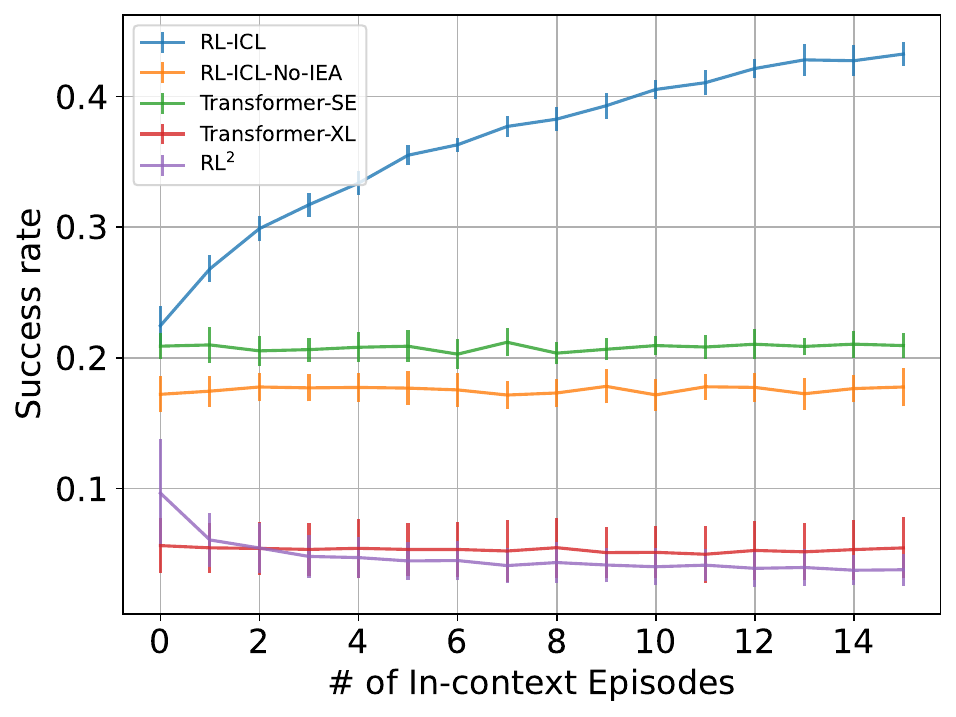}
    \caption{Success Rate}
  \end{subfigure}
  \begin{subfigure}[t]{0.45\columnwidth}
    \includegraphics[width=\textwidth]{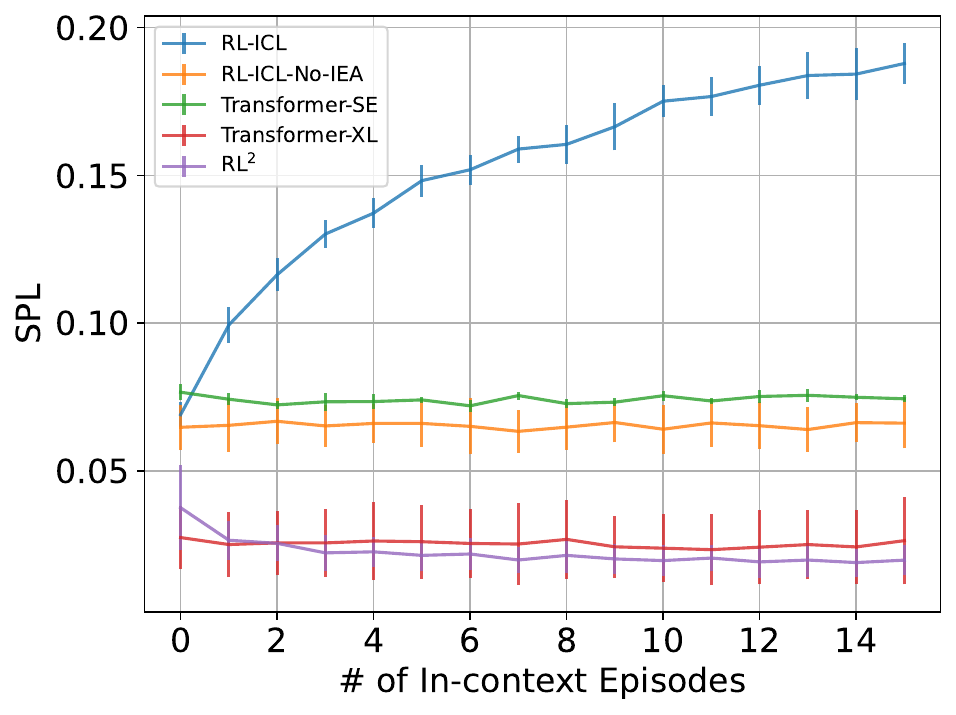}
    \caption{Efficiency}
  \end{subfigure}
  \caption{
    Comparing the in-context learning capability of \method and baselines on \task. The number of episodes in the trial is displayed on the x-axis. The y-axis displays the success or efficiency at that episode count. Agents capable of in-context learning will increase in success and efficiency when encountering more episodes. Each method is run for 3 random seeds and evaluated on 10k distinct sequences. Error bars are standard deviations over trial outcomes between the 3 seeds.}
  \label{fig:icl-results}
\end{figure*}

\section{Experiments}
\label{sec:experiments} 

We first introduce the \Task (\task) task we use to study in-context learning for embodied navigation. Next, we analyze how \method enables in-context learning on this task and outperforms prior work and baselines. We then analyze ablations of \method and analyze its behaviors. We also show \method is capable of few-shot imitation learning. Finally, we show that \method outperforms the methods in \cite{lee2023supervisedpretraininglearnincontext} on the existing Darkroom and Miniworld tasks.

\subsection{\task: \Task}

To evaluate ICL capabilities for embodied agents, we introduce \task, an extension of the existing Object Navigation (\obnav) benchmark. \task assesses an agent's ability to find a sequence of objects in a house while operating from egocentric visual perception. For each object, the agent is randomly placed in a house and must locate and navigate to a specified object category. The agent used is a Fetch robot equipped with a $256 \times 256$ RGB head camera. Additionally, the agent possesses an odometry sensor to measure its relative displacement from the start of the episode. Navigation within the environment is executed through discrete actions: move forward 0.25 meters, turn left or right by 30 degrees, and tilt the camera up and down by 30 degrees. The agent also has a stop action, which ends the episode.

\task uses scenes from the Habitat Synthetic Scenes Dataset (HSSD)~\citep{khanna2023habitat} along with a subset of the YCB object dataset~\citep{calli2015ycb} containing 20 objects types. 
Note that \task requires navigating to \emph{small} objects unlike other \obnav variants that use large receptacles as goals \citep{habitat-challenge-2021}. This allows us to increase the dataset diversity by sampling objects randomly in the environment, unlike \obnav, where the receptacles are fixed parts of the scanned meshes. The random sampling also precludes the agent from using priors over object placements in scenes, forcing it to rely on the experience in its context.

\task defines a \emph{trial} as a sequence of episodes within a fixed home layout, where a home layout is defined by a combination of a floorplan, a furniture layout and set of object placements. A home layout contains on an average 22 objects where multiple object instances may be of the target category. Within an episode in the trial, a target object category is randomly selected and the agent is randomly placed in the house. The episode is successful if the agent calls the stop action within 2 meters of the object, with at least 10 pixels of the object in the current view. If the object is not found within $500$ steps the episode counts as a failure.

We evaluate the agents on unseen scenes from HSSD and report the success rate (SR) and Success-weighted by Path Length (SPL) metrics~\citep{anderson2018evaluation}. Specifically, we look at the SR and SPL of an agent as it accumulates more episodes in-context. Ideally, with more in-context episodes within a home layout, it should be more adept at finding objects and its SR and SPL should improve. See \Cref{sec:task-details} for further details on the \task.

\begin{figure*}[t]
  \centering
  \begin{subfigure}[t]{0.32\columnwidth}
    \includegraphics[width=\textwidth]{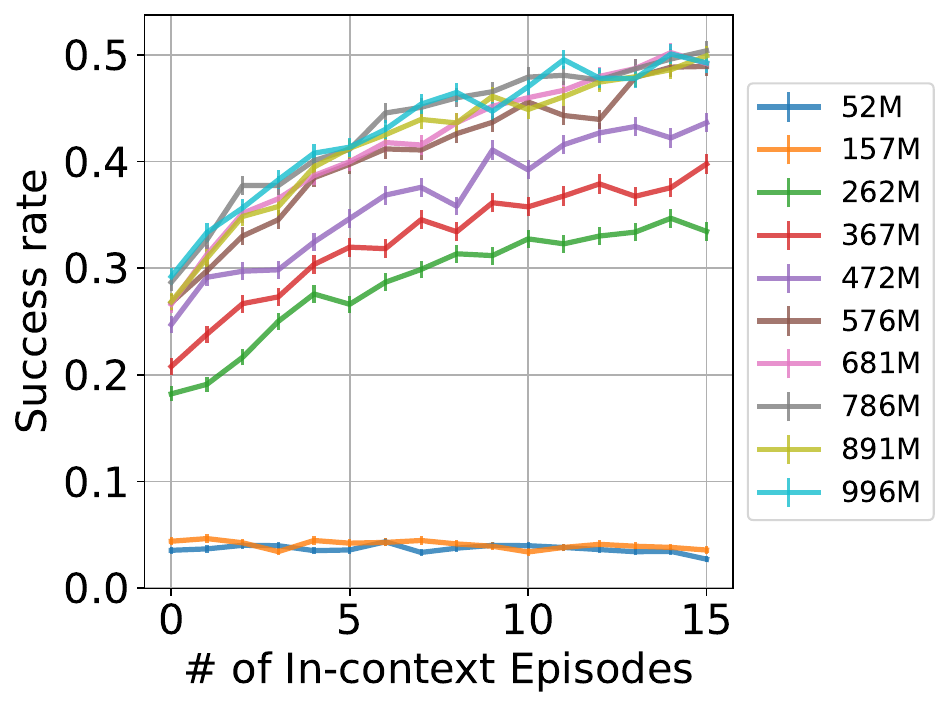}
    \caption{Learning vs. ICL}
    \label{fig:icl-vs-updates} 
  \end{subfigure}
  \begin{subfigure}[t]{0.32\columnwidth}
    \includegraphics[width=\textwidth]{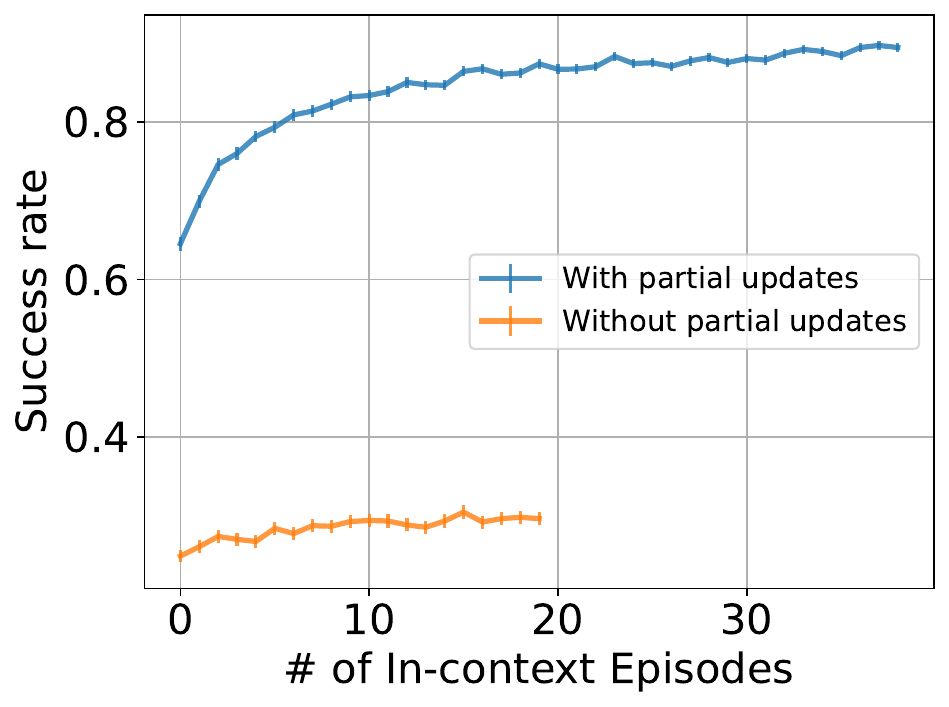}
    \caption{Update Scheme Ablation}
    \label{fig:updates} 
  \end{subfigure}
  \begin{subfigure}[t]{0.32\columnwidth}
    \includegraphics[width=\textwidth]{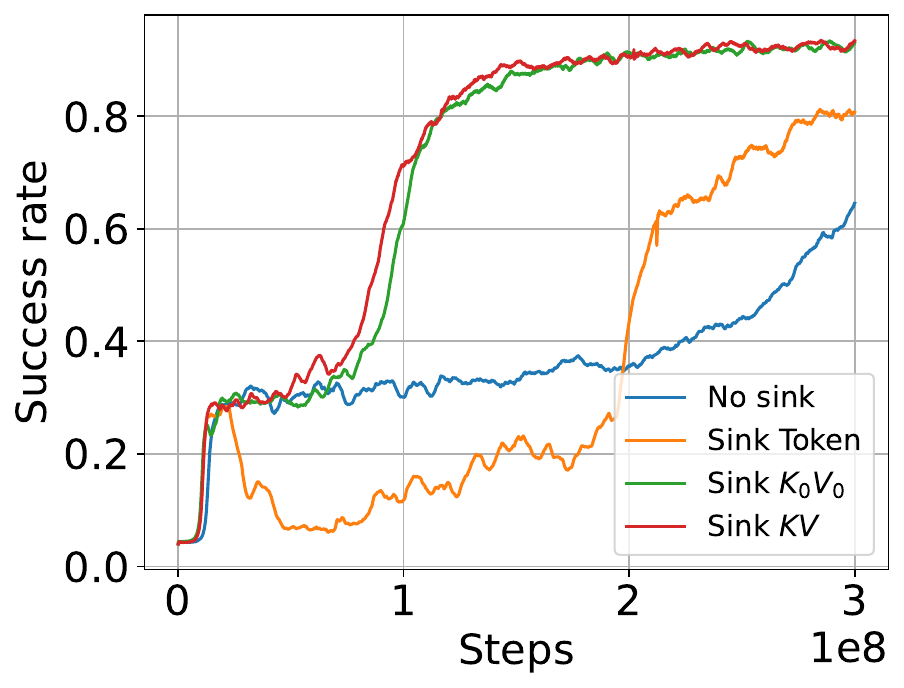}
    \caption{Attention Sink Ablation}  
    \label{fig:sink-kv-learning-curve}
  \end{subfigure}
  \caption{
    Analyzing \method ICL capabilities.
    \cref{fig:icl-vs-updates} shows increased RL training results in agents that have a higher base success and stronger ICL capabilities with error bars giving standard error on the evaluation episodes.
    \cref{fig:updates} shows the partial updates are important in \method. 
    \cref{fig:sink-kv-learning-curve} shows \sinkv is important for learning speed and stability. 
    The results in Fig.\ref{fig:updates},\ref{fig:sink-kv-learning-curve} use the smaller ReplicaCAD scenes for easier analysis and thus have higher overall success rates. 
    These results performed on the easier ReplicaCAD scenes to save compute, so the numbers are higher overall.
  }
  \label{fig:analysis}
  \vspace*{-15pt}
\end{figure*}

\subsection{In-Context Learning on \task}
\label{sec:icl_on_task}

In this section, we compare the ability of \method and baselines to in-context learn in a new home layout. We compare \method to the following baselines:

\begin{itemize}[itemsep=2pt,topsep=0pt,parsep=0pt,partopsep=0pt,parsep=0pt,leftmargin=*]
  \item \textbf{\rls} \cite{duan2016rl}: Use an LSTM and keep the hidden state between trial episodes.
  \item \textbf{\trxl (TrXL)} \cite{dai2019transformerxl}: Use Transformer-XL and updates the constant-size memory recurrently. This is the model used in \cite{team2023human} trained in our setting. Following~\cite{team2023human} we use PreNorm \citep{gtrxl} and gating in the feedforward layers \citep{shazeer2020glu}. 
  \item \textbf{\ieafmethod}: \method without Inter-Episode Attention (IEA). Everything else, including the update scheme is the same as \method.
  \item \textbf{\semethod}: A transformer-based policy operating over only a single episode (SE) and without the update schemes from \method. 
\end{itemize}

All baselines are trained for 500M steps using a distributed version of PPO~\citep{wijmans2019dd}. Methods that utilize multi-episode context are trained with a context length of 4k, and use 8k context length during inference (unless mentioned otherwise, e.g. in our long-context experiments). The results in \Cref{fig:icl-results} demonstrate \method achieves better performance than baselines on 8k steps of ICL, achieving $ 43\%$ success rate v.s.  $ 22 \%$ success rate achieved by the closest performing baseline (\semethod).

Additionally, \method is able to effectively adapt to new home layouts throughout the course of the trial. In the first episode of the trial, transformer-based baseline methods attain a similar base performance of around $ 20 \%$ success rate. However, as more episodes arrive, the performance of \method increases. The recurrent models, Transformer-XL and \rls, have lower base performance at $10\%$ success rate and show no in-context learning. 
The performance of \rls degrades with more in-context episodes, which is aligned with the inability of the LSTM to model long sequences.

After 15 episodes of in-context experience, the success rate of \method increases from $ 23\%$ to $ 43\%$.  
The baselines do not possess this same ICL ability and maintain constant performance with subsequent in-context episodes.
\method also in-context learns to navigate faster to objects, as measured by the  gap in SPL.
As the trial progresses, the agent is able to more efficiently navigate to objects in the house with the SPL of \method increasing from $ 0.07$ to $ 0.188$. 
The baselines are unable to improve efficiency in-context and maintain a SPL of $ 0.025$ to $ 0.075$ throughout the entire trial.

\vspace{-5pt}
\subsection{\method Ablations and Analysis}
\label{sec:ablation}
\vspace{-5pt}

We demonstrate that the partial updates in \method and \sinkv are crucial to learning with RL over long context windows and acquiring ICL capabilities. We run these ablations in the smaller ReplicaCAD~\citep{szot2021habitat} scenes to make methods faster to train, but other details of the task remain the same.
We then show that ICL emerges later in the training and the context length in \method can be even further increased.

\textbf{No Partial Updates.} Firstly, we remove the partial updates in \method and find that it performs poorly (\Cref{fig:updates}), achieving $ 40 \%$ lesser SR at the first episode. This model also shows little ICL abilities with the SR only increasing $ 5 \% $ by the end of the trial versus a $ 25\%$ increase when using partial updates. 

\textbf{\sinkv.} Next, we demonstrate that using \sinkv is necessary for sample-efficient in-context RL learning. We trained the model on ReplicaCAD with and without attention sinks. The learning curves in \Cref{fig:sink-kv-learning-curve} shows that learning is more stable and faster with \sinkv which achieves 90\% success rate at 200M steps. It also shows that \sinkv performs similar to Softmax One \citep{off_by_one_softmax}, referred to as Sink $K_0V_0$. 
Without attention sink mechanisms, learning is slow and achieves less than 40\% success rate after 200M steps and reaching 64\% at 300M steps. Using sink token \citep{xiao2023efficient}, the training becomes unstable, achieving 40\% success rate at 200M steps training and reaching 80\% success rate at 300M steps. The details of the different sink attentions, their implementations and how the attention heads use the \sinkv can be found in \Cref{sec:app-sink-kv}.

\textbf{Training Steps v.s. ICL Abilities.} We find that \method only acquires ICL capabilities after sufficient RL training. As demonstrated in \Cref{fig:icl-vs-updates}, the agent is only capable of ICL after 157M steps of training. Models trained for 52M and 157M remain at constant success with more in-context experience. Further training does more than just increase the base agent performance in the first episode of the trial. 
From 262M steps to 367M steps, the agent base performance increases by $ 2\%$, yet the performance after 15 episodes of ICL performance increases $ 10 \%$.
This demonstrates that further training is not only improving the base capabilities of the agent to find objects, but also improving the agent's ability to utilize its context across long trials spanning many episodes.

\textbf{Context length generalization.} Next, we push the abilities of \method to in-context learn over contexts much larger than what is seen during training. In this experiment, we evaluate \method model, trained with 4k context length, on 32k steps of experience, which is enough to fit 80 episode trials in context. Assuming that the simulator is operating at 10Hz, this is almost 1 hour of agent experience within the context window. Note that for this experiment, we use our best checkpoint, which is trained for 1B steps. The results demonstrate that \method can generalize to contexts 8$\times$ larger at inference. \Cref{fig:ctx-gen} shows \method is able to further increase the success rate to over $ 55\% $ after 80 in context episodes and consistently maintains performance above $ 50 \% $ after 20 in-context episodes.

\textbf{64k steps trials.} Finally, we investigate scaling \textit{training} \method with 64k context length. We use the same hyperparameters as \cref{sec:icl_on_task}, but increase the number of partial updates per rollout such that the policy is updated every 256 steps, the same number of steps used in \method. \cref{fig:64k-train-succ} shows that the model can in-context learn over 175 episode and continue to improve success rate. More details are available at \Cref{sec:app-64k-train-eval}.

In \Cref{sec:per-obj} we analyze the performance of \method per object type. In \Cref{sec:app-what-agent-see} we qualitatively analyze what the agent attends to in successful and failure episodes. Finally, in \Cref{sec:shuffle-abl}, we show that not shuffling episodes in the context during training leads to worse performance.

\vspace{-5pt}
\subsection{Emergent Few-Shot Imitation Learning}
\vspace{-5pt}
\begin{figure*}[t]
  \centering
  \begin{subfigure}[t]{0.32\columnwidth}
    \includegraphics[width=\textwidth]{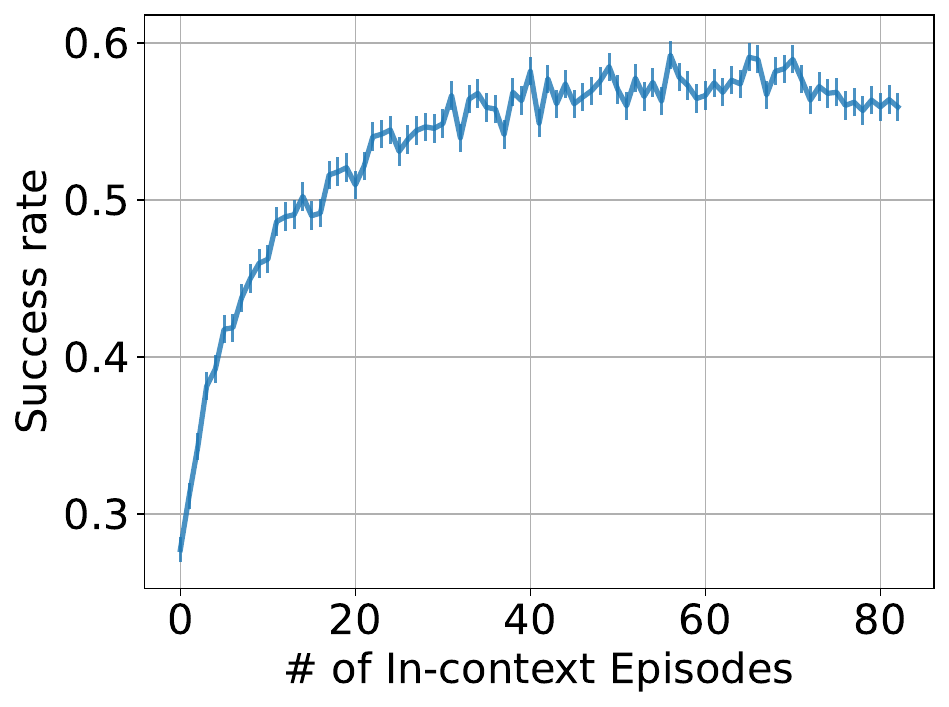}
    \caption{Context Length Generalization}
    \label{fig:ctx-gen}
  \end{subfigure}
  \begin{subfigure}[t]{0.32\columnwidth}
    \includegraphics[width=\textwidth]{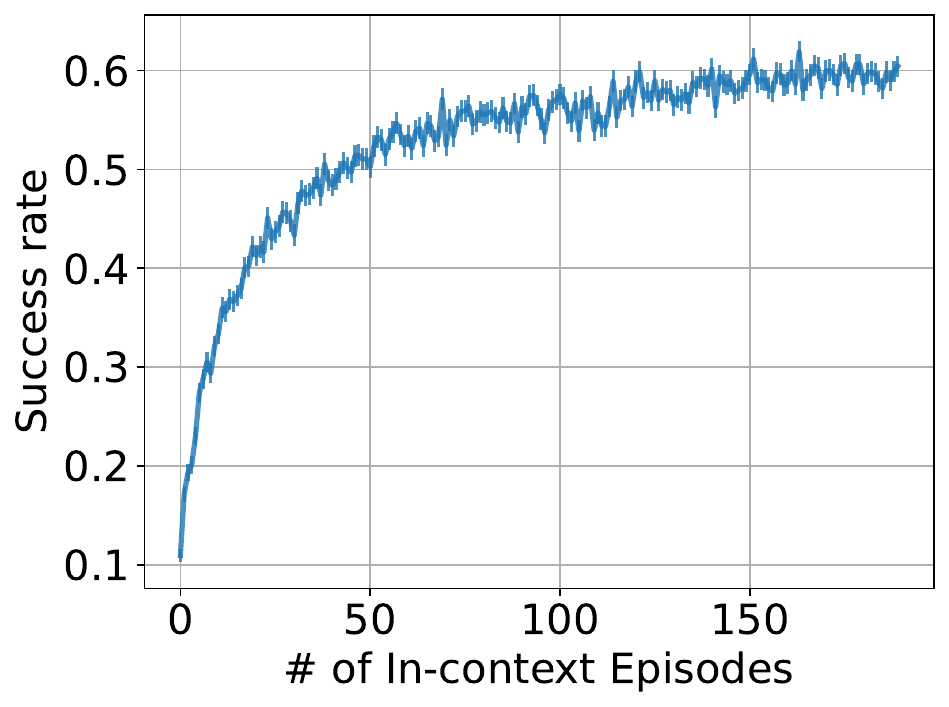}
    \caption{64k Train Context Length}
    \label{fig:64k-train-succ}
  \end{subfigure}
  \begin{subfigure}[t]{0.32\columnwidth}
    \includegraphics[width=\textwidth]{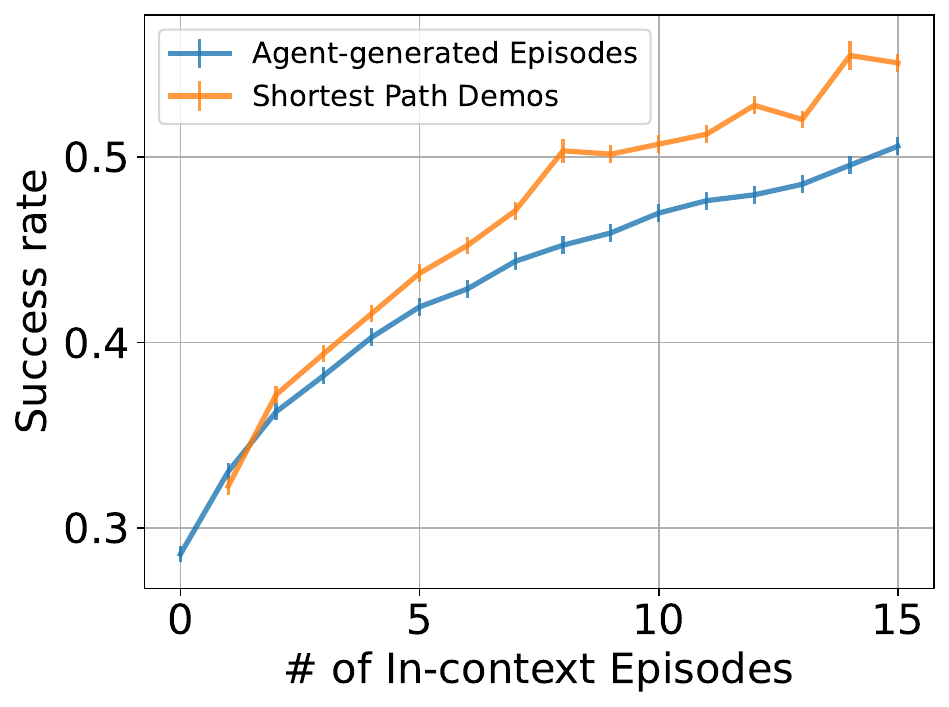}
    \caption{Few-Shot Imitation Learning}
    \label{fig:fsil}
  \end{subfigure}
  \vspace{-5pt}
  \caption{(a) \method trained with context length 4k generalizes to operating at 32k steps of in context experience in a new home layout. (b) \method trained at 64k context length shows ICL abilities over 175 episodes. (c) \method can do few-shot imitation learning despite not training for it. The error bars represent the standard error.}
  \vspace{-15pt}
\end{figure*}

In addition to learning in-context from self-generated experience in an environment, \method can also use its context to learn from demonstrations provided by an external agent or expert, despite never being trained on demonstrations and only learning from self-generated experience. 
We consider the setting of few-shot imitation learning \citep{duan2017one,wang2020generalizing} where an agent is given a set of trajectories $ \left\{ \tau_1, \dots,  \tau_N \right\} $ demonstrating behavior reaching desired goals $\left\{ g_1, \dots,  g_N \right\} $. The agent must then achieve a new $ g_{N+1}$ in new environment configurations using these demonstrations. 
\method is able to few-shot imitation learn by taking the expert demonstration as input via the context. Specifically, we generate $ N$ expert shortest path trajectories navigating to random objects from random start positions in an unseen home layout.
The success rate of these demos is around $ 80\%$ due to object occlusions hindering the shortest path agent from viewing the target object which is required for success.
These $ N$ trajectories are inserted into the context of \method and the agent is instructed to navigate to a new object in the environment. 

In \Cref{fig:fsil} we show that \method can utilize these expert demonstrations despite never seeing such shortest paths during training. \Cref{fig:fsil} shows the success rate of \method in a single episode after conditioning on some number of shortest path demonstrations. More demonstrations cover more of the house and the agent is able to improve navigation success.
We also compare to the success rate of an agent that has $ N$ episodes of experience in the house as opposed to $ N$ demonstrations. Using the demonstrations results in better performance with $ 5\%$ higher success rate for $ N=16$. 

\vspace{-5pt}
\subsection{Darkroom and Miniworld}
\label{sec:darkroom} 
\vspace{-5pt}

\begin{figure*}[t]
  \centering
  \begin{subfigure}[t]{0.32\columnwidth}
    \includegraphics[width=\textwidth]{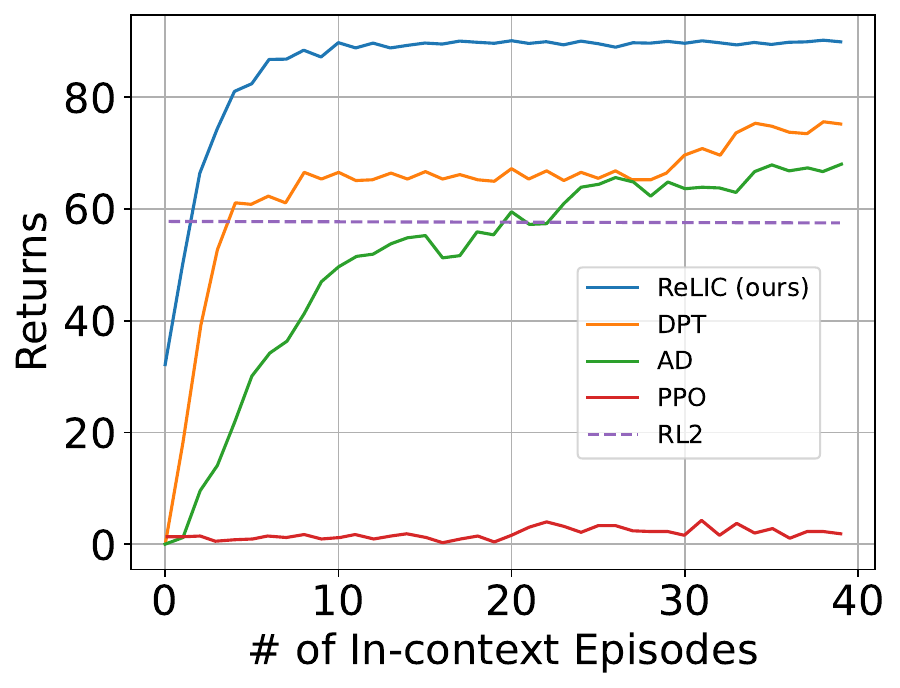}
    \caption{Darkroom}
    \label{fig:dark_result}
  \end{subfigure}
  \begin{subfigure}[t]{0.32\columnwidth}
    \includegraphics[width=\textwidth]{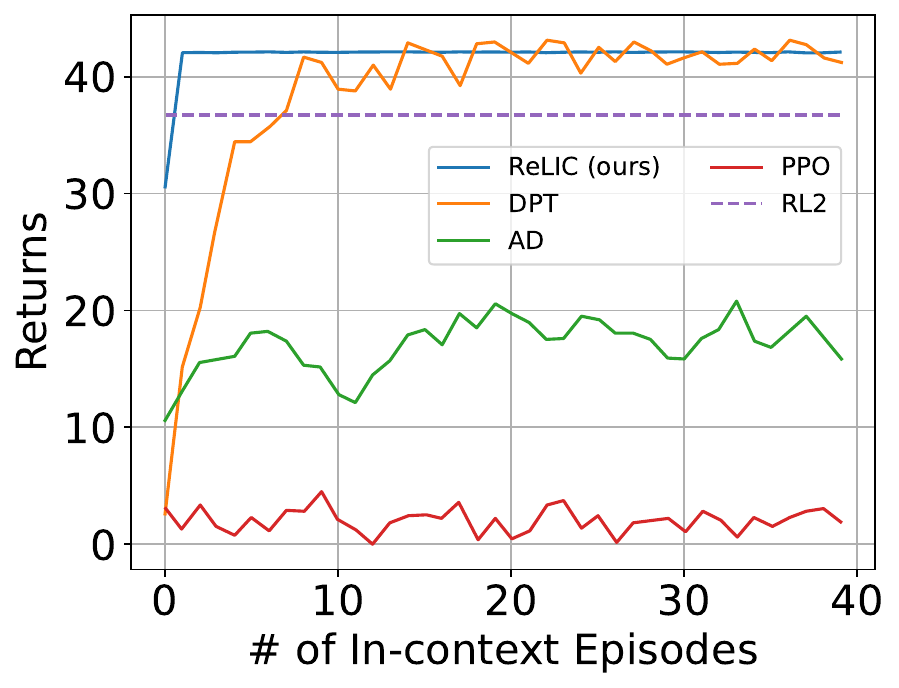}
    \caption{Miniworld}
    \label{fig:mini_result}
  \end{subfigure}
  \vspace{-5pt}
  \caption{
    ICL comparison of \method and baselines in the Darkroom and Miniworld tasks. 
    \method has a higher base performance and adapts to new tasks with less experience. The baselines numbers are obtained from Figures 4b,d of \cite{lee2023supervisedpretraininglearnincontext}.
    } 
  \label{fig:dark_n_mini_result}
  \vspace{-10pt}
\end{figure*}

In this section, we evaluate \method on the Darkroom \citep{zintgraf2020varibadgoodmethodbayesadaptive} and Miniworld~\citep{gym_miniworld} environments and compare to the results from \cite{lee2023supervisedpretraininglearnincontext} to provide a comparison with existing baselines on these simpler benchmarks.
We directly take the numbers from \cite{lee2023supervisedpretraininglearnincontext} which include Decision-Pretrained Transformer (DPT), a supervised pretraining method for in-context meta-RL, Algorithm Distillation (AD) \citep{laskin2022context}, Proximal Policy Optimization (PPO) and \rls. 
\method is trained with context length 512, which fits 10 Miniworld and 5 Darkroom episodes. 
Policies are evaluated with 40 in-context episodes. 
Full details are in \Cref{sec:dark_n_min_hyperparams}.

\textbf{Darkroom.} \Cref{fig:dark_result} shows that \method outperforms all previous methods in the Darkroom task. Specifically, \method achieves base performance of 32 while the other methods have base performance lower than 2 returns. \method reaches 89 returns after 10 in-context episodes which is higher than 75 achieved by DPT after 39 in-context episodes.

\textbf{Miniworld.} \method has a higher base performance of 30 episode return compared to the best base performance of 10 as shown in \Cref{fig:mini_result}. It quickly reaches 42 returns after just 2 in-context episodes while DPT reaches the same result after 14 in-context episodes. \method also shows stable performance as the number of episodes increase compared to DPT which shows oscillation in the performance. 
In \Cref{sec:miniworld_abls}, we also show the importance of \sinkv and partial updates in both tasks.

\vspace{-10pt}
\section{Conclusion and Limitations}
  \vspace{-10pt}

\looseness=-1 The ability of an agent to rapidly adapt to new environments is crucial for successful Embodied AI tasks. We introduced \method, an in-context RL method that enables the agent to adapt to new environments by in-context learning with up to 64k environment interactions and visual observations. We studied the two main components of \method: \textit{partial updates} and the \textit{\sinkv} and showed both are necessary for achieving such in-context learning. 
We showed that \method results in significantly better performance on a challenging long-sequence visual task compared to the baselines.

Limitations of the approach are that we found for ICL to emerge, it requires a diverse training dataset on which the model can not overfit. There is no incentive for the model to learn to use the context if it can overfit the task. We were able to address that in the dataset generation by creating different object arrangements for each scene which made it challenging for the model to memorize the objects arrangements.
Another is that our study only focuses on several environments.
Future work can explore this same study in more varied environments such as a mobile manipulation task where an agent needs to rearrange objects throughout the scene.
Finally, \method requires large amounts of RL training to obtain in-context learning capabilities.

\section{Acknowledgements}

This work was supported in part by NSF, ONR YIP, and ARO PECASE. The views and conclusions contained herein are those of the authors and should not be interpreted as necessarily representing the official policies or endorsements, either expressed or implied, of the U.S. Government, or any sponsor.

\bibliography{iclr2025_conference}
\bibliographystyle{iclr2025_conference}

\appendix
\newpage

\section{Additional \task Details}
\label{sec:task-details}

\begin{figure*}[t]
  \centering
  \begin{subfigure}[t]{0.3\columnwidth}
    \includegraphics[width=\textwidth]{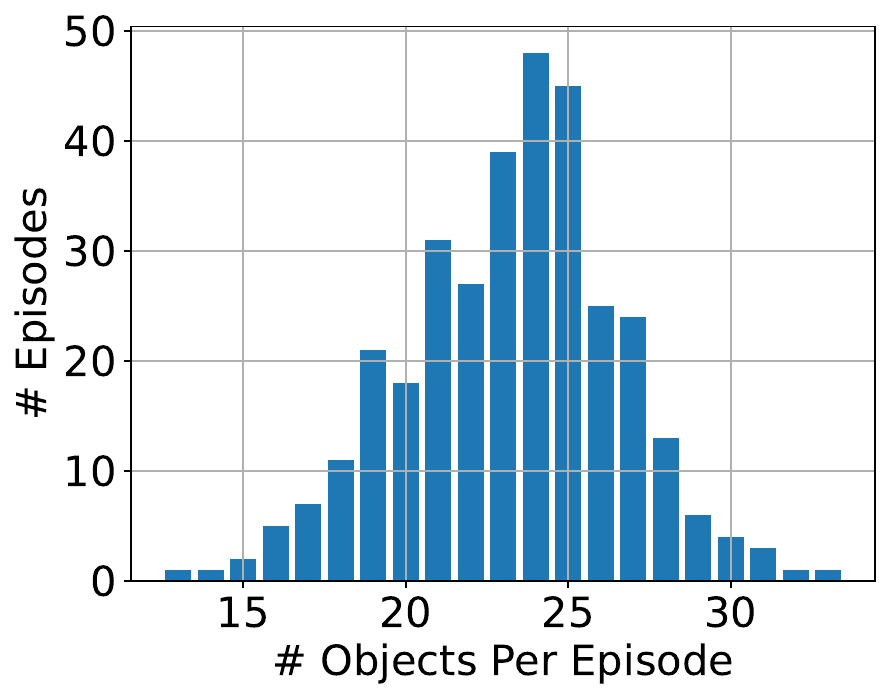}
    \caption{\# Objects Per Episode}
  \end{subfigure}
  \begin{subfigure}[t]{0.3\columnwidth}
    \includegraphics[width=\textwidth]{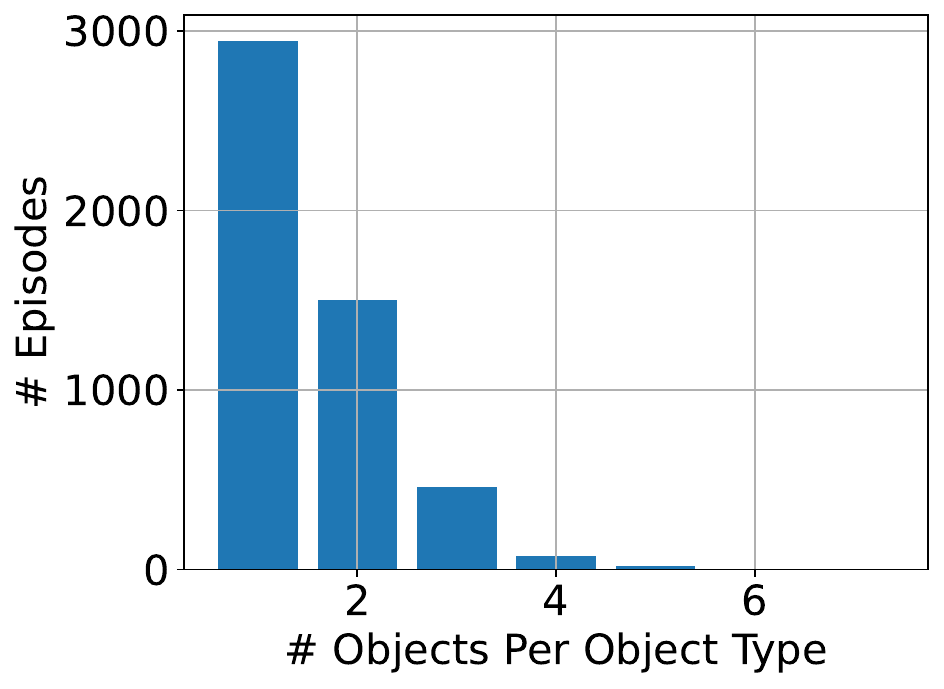}
    \caption{\# Objects Per Object Type}
  \end{subfigure}
  \begin{subfigure}[t]{0.3\columnwidth}
    \includegraphics[width=\textwidth]{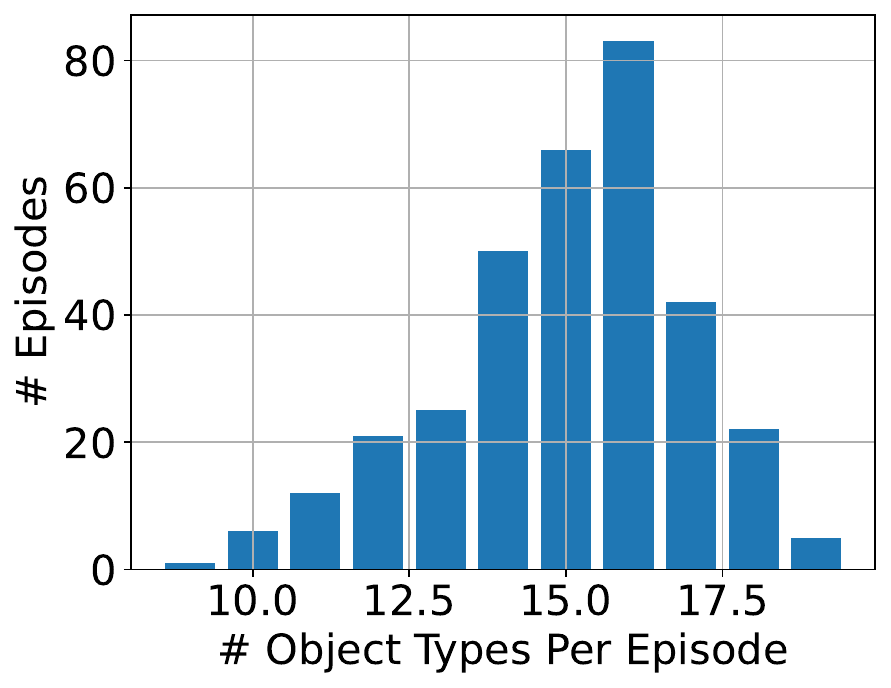}
    \caption{\# Object Types Per Episode}
  \end{subfigure}
  \caption{
    The distribution of the objects and object types in the data.
  }
  \label{fig:app-data-dist}
\end{figure*}

The \task is a small object navigation task. We use the same training (37) and validation (12) scenes as \cite{yenamandra2023homerobot}. The data is generated by randomly placing objects from the 20 object types, a subset of the YCB \cite{calli2015ycb} dataset, on random receptacles. The data is generated by sampling between 30 to 40 object instances and placing them on receptacles, and subsequent filtering of the objects that are not reachable by the agent. The filtering is done by placing the agent in front of the object and evaluating whether the agent can meet the success criteria. If the agent can meet the critera, we retain the object. Otherwise, we discard the object. The distribution of the objects and object types are in \Cref{fig:app-data-dist}.

The reward function is defined as follows:
\begin{itemize}[itemsep=2pt,topsep=0pt,parsep=0pt,partopsep=0pt,parsep=0pt,leftmargin=*]
    \item Change in geodesic distance to the closest object $r_d=-\Delta{d}$ where $d$ is the geodesic distance to the closest object. The closest object can change across the episode.
    \item Slack reward of $-0.001$.
    \item Succes reward of $2$.
\end{itemize}

The episode is considered a success if the agent selects the \textit{Stop} action while it is within 2 meters of an instance of the target type and has 10 pixels of this instance in the view.

\section{Additional Method Details}

\subsection{Model training}

In this section, we discuss the training setup for \method experiment. 

\textbf{The workers and the batch size.} We use 20 environment workers per GPU. Since we use 4 GPUs in parallel, there are 80 environment workers in total. The micro batch size is 1 and we accumulate the gradient for 10 micro batches on the 4 GPUs which makes the effective batch size 40.

\textbf{RL algorithm.} We use PPO \cite{schulman2017proximal} to train the model with $\gamma=0.99$, $\tau=0.95$, entropy coefficient of 0.1 and value loss coefficient of 0.5.

\textbf{Optimizer.} We use Adam \cite{kingma2014adam} optimizer to learn the parameters.

\textbf{Learning rate schedule.} We use learning rate warm up in the first 100,000 environment interactions. The learning rate starts with $LR_0=2e-7$ and reaches $LR=2e-4$ at the end of the warm up. Cosine decay \cite{loshchilov2017sgdr} is used after the warm-up to decay the learning rate to $0$ after 1B environment interactions. 

\textbf{Precision.} We use FP16 precision for the visual encoder and keep the other components of the model as FP32.

\textbf{Rollout Storage.} The rollout storage size is 4096. We store the observations and the visual embeddings in rollout in low-precision storage, specifically in FP16 precision.

\textbf{Regularization.} We follow \cite{reed2022generalist} in using depth dropout \cite{huang2016deep} with value 0.1 as regularization technique. We also shuffle the in-context episodes after each partial updates.

\textbf{Hardware Resources and Training Time.} The model is trained for 1B steps on 4x Nvida A40 for 12 days.

\subsection{Hyperparameters}
\label{sec:hyperparams} 

We list the hyperparameters for the different experiments discussed in section \cref{sec:icl_on_task}.

\textbf{{\method}}: The hyperparameters used in \method can be found in \Cref{tab:method-hparam} and the hyperparameters of the transformer model used in the training can be found in \Cref{tab:transformer-hparam}. 

\textbf{\rls}: For implementing \rls, we build on the default PPO-GRU baseline parameters in Habitat 3.0 \cite{puig2023habitat3}. We set the number of PPO update steps to 256, and the hidden size of the GRU to 512. The scene is changed every 4096 steps during training, and the hidden state is reset to zeros after every scene change.

\textbf{\ieafmethod}: We use the same model and hyperparameters as \method. The only difference is that we set the attention mask to restrict the token to only access other tokens within the same episode.

\textbf{\semethod}: We use the same model and hyperparameters as \method. However, we limit the training sequence to a fixed size 385 old observations + 256 new observations. The choice of the old number of observations is made such that we never truncate an episode which is at most 500 steps. The attention mask is set to restrict the tokens to only access other tokens in the same episode.

\textbf{\trxl (TrXL)} \cite{dai2019transformerxl}: Use Transformer-XL and update the constant-size memory recurrently. We follow \cite{team2023human} in that we use PreNorm \cite{gtrxl} and use gating in the feedforward layers \cite{shazeer2020glu}. We experiment with two values for the memory size, 256 and 1024, using TrXL without gating and found that the model is able to learn with 256 memory but is unstable with 1024 memory. We use 256 memory size which gives the agent context of size $L \times N_m=4\times 256=1024$ where $L$ is the number of layers. Except for the memory, we use the same number of layers and heads and the same hidden dimensions as \method.

\begin{table}[!h]
    \centering
    \medskip
  \begin{tabular}{ll}
    \toprule
    Hyperparameter     & Value \\
    \midrule
    \# Layers & 4 \\
    \# Heads & 8 \\
    Hidden dimensions & 256 \\
    MLP Hidden dimensions & 1024 \\
    \# Sink-KV & 1 \\
    Attention sink & Sink $KV_0$ \\
    Episode index encoding & RoPE \cite{su2023roformer} \\
    Within-episode position encoding & Learnable \\
    Activation & GeLU \cite{shazeer2020glu} \\
    Rollout size & 4096 \\
    total \# updates per rollout & 16 \\
    \# partial updates & 15 \\
    \# full updates & 1 \\
    \bottomrule
  \end{tabular}
  \centering
  \caption{\method and baseline hyperparameters}
  \label{tab:method-hparam}
  \label{tab:transformer-hparam}
\end{table}

\subsection{Darkroom and Miniworld Hyperparameters}
\label{sec:dark_n_min_hyperparams}

We use smaller transformer for these two tasks described in \cref{tab:tiny-transformer-hparam}. The \method hyperparameters are provided in \cref{tab:darkroom-miniworld-hparam}. For the visual encoder, we use the CNN model used in \cite{lee2023supervisedpretraininglearnincontext} and train it from scratch. The other hyperparameters are the same as described in \cref{sec:hyperparams}.

\begin{table}[!htb]
  \centering
  \begin{tabular}{ll}
    \toprule
    Hyperparameter     & Value \\
    \midrule
    \# Layers & 2 \\
    \# Heads & 8 \\
    Hidden dimensions & 64 \\
    MLP Hidden dimensions & 256 \\
    \# Sink-KV & 1 \\
    Attention sink & Sink $K_0V_0$ \\
    Episode index encoding & RoPE \cite{su2023roformer} \\
    Within-episode position encoding & Learnable \\
    Activation & GeLU \cite{shazeer2020glu} \\
    Rollout size & 512 \\
    \# updates per rollout & 4 (Darkroom), 2 (Miniworld) \\
    \# partial updates & 3 (Darkroom), 1 (Miniworld) \\
    \# full updates & 1  \\
    \bottomrule
  \end{tabular}
  \caption{Hyperparameters for \method and baselines in Miniworld and Darkroom.}
  \label{tab:tiny-transformer-hparam}
  \label{tab:darkroom-miniworld-hparam}
\end{table}

\section{More Experiments}

\subsection{\method per Object Type}
\label{sec:per-obj} 
\begin{figure*}[t]
  \centering
  \begin{subfigure}[t]{0.45\columnwidth}
    \includegraphics[width=\textwidth]{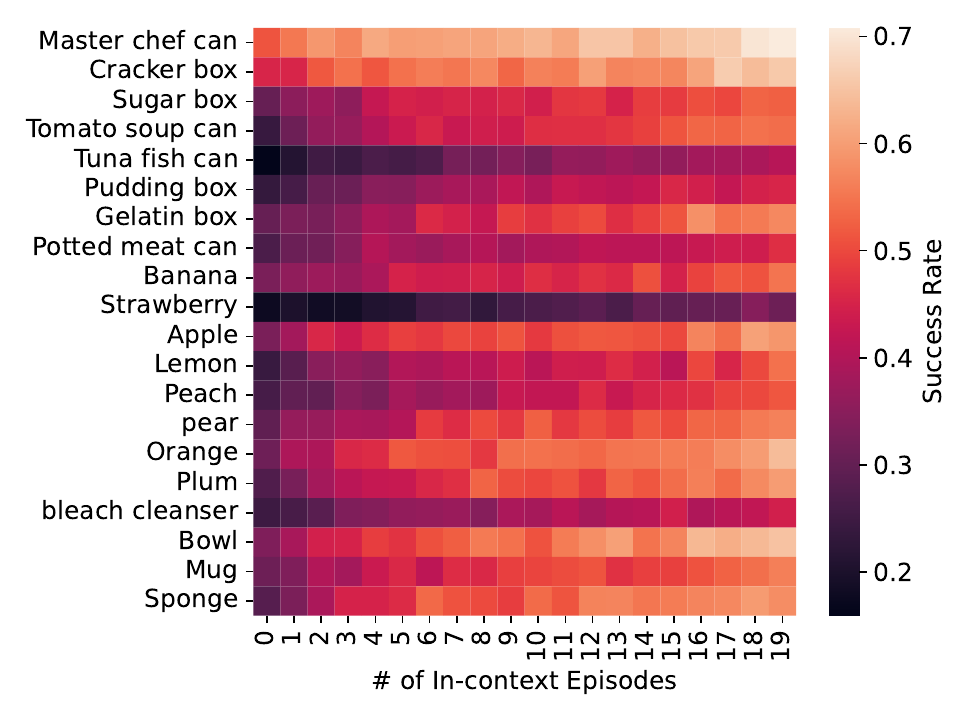}
    \caption{Fixed Object Type Per Trial}
    \label{fig:obj-analysis} 
  \end{subfigure}
  \begin{subfigure}[t]{0.45\columnwidth}
    \includegraphics[width=\textwidth]{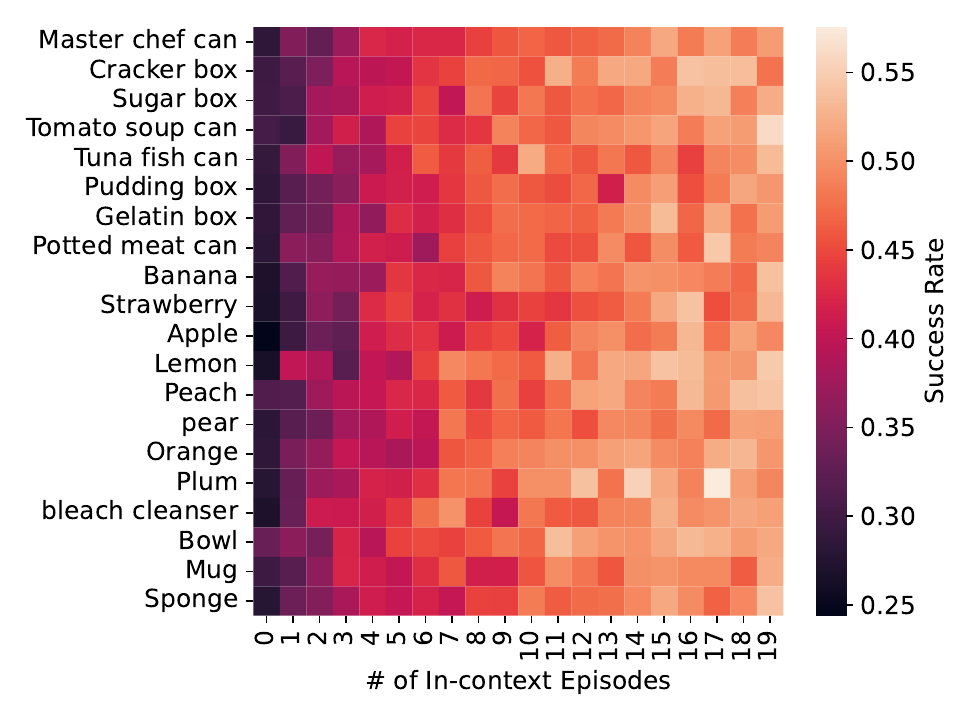}
    \caption{\method per Object Type}
    \label{fig:obj-first-analysis} 
  \end{subfigure}
  \caption{
    Analysis of how \method learns to navigate to particular object types through ICL. 
    (a) compares the number of consecutive episodes within a trial an object appears v.s. the success rate. The agent becomes more capable at navigating to that object type for subsequent episodes. 
    (b) shows the episode index within the trial that the object first appears v.s. the average success rate for different objects. As the agent acquires more experience in-context, it can proficiently navigate to any object type.
  }
  \label{fig:att}
\end{figure*}

In \Cref{fig:obj-analysis}, we analyze the ICL performance of \method per object type. Specifically, we specify the same object type target for the agent repeatedly for 19 episodes. Similar to the main experiments, the agent is randomly spawned in the house. As \Cref{fig:obj-analysis} illustrates, \method becomes more capable at navigating to all object types in subsequent episodes. The agent is good at adapting to finding some objects such as bowls, cracker box, and apples. Other objects, such as strawberry and tuna fish can, remain difficult. 
In \Cref{fig:obj-first-analysis}, we show that with 19 episodes of ICL, the agent is can reliably navigate to any object type in the house despite having different object types as target in the context. This demonstrates the agent is able to utilize information about other object targets from the context.

\begin{figure*}[t]
  \centering
  \begin{subfigure}[t]{0.45\columnwidth}
    \includegraphics[width=\textwidth]{figures/results/64ksteps_training_inference_custom_pddl_task_success.pdf}
    \caption{Success Rate}
  \end{subfigure}
  \begin{subfigure}[t]{0.45\columnwidth}
    \includegraphics[width=\textwidth]{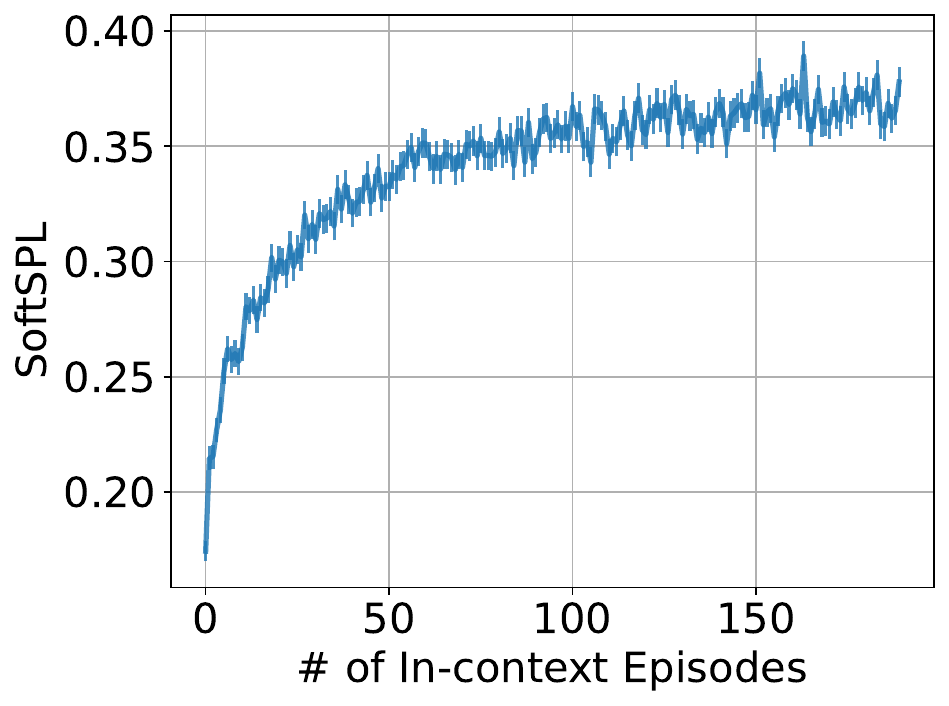}
    \caption{Efficiency}
  \end{subfigure}
  \caption{
    The success and efficiency of training and evaluating \method with 64k context length.
  }
  \label{fig:app-64k-training}
\end{figure*}

\subsection{Analyzing Attention Scores}
\label{sec:app-what-agent-see} 

\begin{figure*}[t!]
  \centering
  \begin{subfigure}[t]{0.24\columnwidth}
    \includegraphics[height=\textwidth]{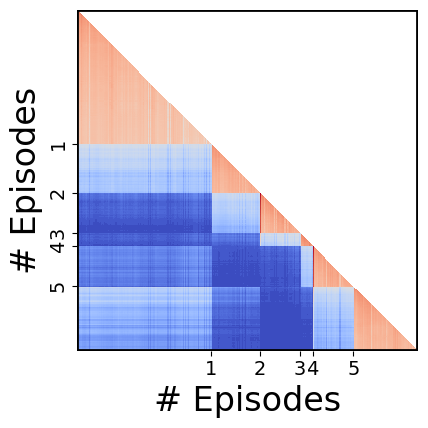}
    \caption{Intra-episode attn}
    \label{fig:attn_score_within_eps} 
  \end{subfigure}
  \begin{subfigure}[t]{0.24\columnwidth}
    \includegraphics[height=\textwidth]{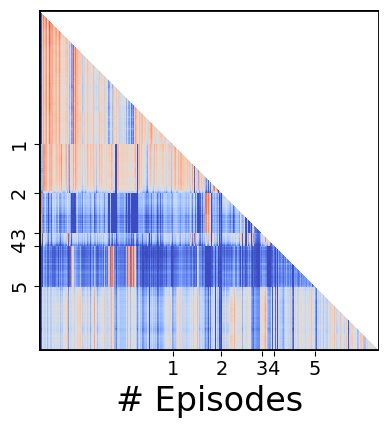}
    \caption{Inter-episodes attn}
    \label{fig:attn_score_iea} 
  \end{subfigure}
  \begin{subfigure}[t]{0.24\columnwidth}
    \includegraphics[height=\textwidth]{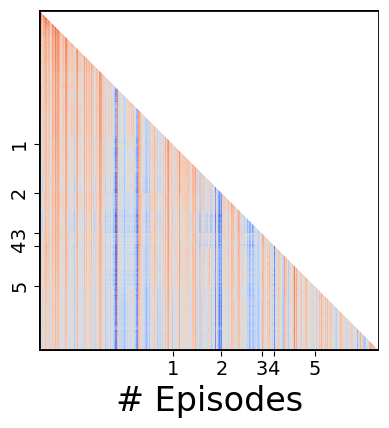}
    \caption{Episode-invariant attn}
    \label{fig:attn_score_eps_invariant} 
  \end{subfigure}
  \begin{subfigure}[t]{0.24\columnwidth}
    \includegraphics[height=\textwidth]{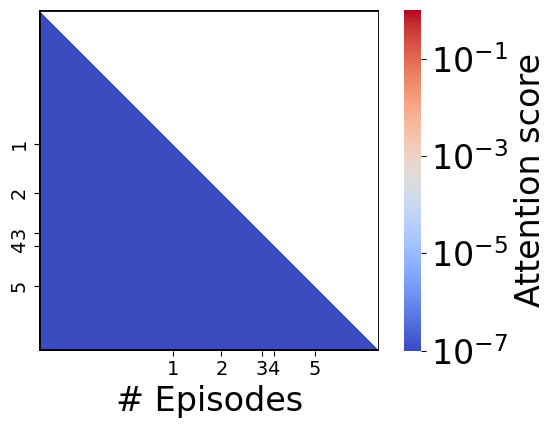}
    \caption{Zero attn}
    \label{fig:attn_score_0} 
  \end{subfigure}
  \vspace{-5pt}
  \caption{
    Attention scores patterns of a sequence with 1024 steps. We found 4 attention patterns in the heads of a trained policy: (a) Intra-episode attention where the attention head assigns high score to the running episode, (b) Inter-episode attention pattern where the attention head assigns high score to the context, without being constrained to the running episode, (c) the episode-invariant pattern where the attention head attends to the same tokens regardless of the episode structure in the context, and (d) the zero attention pattern where the attention head assign all attention scores to the \sinkv.
  }
  \vspace{-15pt}
  \label{fig:attn_score_patts}
\end{figure*}

In this section, we show that the agent is able to utilize the in-context information by inspecting the attention scores patterns in the attention heads. We generate the data by letting the agent interact with an unseen environment for 19 episodes which produced a sequence of 2455 steps. A random object type is selected as a target in each episode. By inspecting the attention scores of the attention heads, we found 4 patterns shown in \Cref{fig:attn_score_patts}. 

\begin{itemize}[itemsep=2pt,topsep=0pt,parsep=0pt,partopsep=0pt,parsep=0pt,leftmargin=*]
  \item \textbf{Intra-episode attention}: In this pattern, the agent attends only to the running episode, \Cref{fig:attn_score_within_eps}.
  \item \textbf{Inter-episodes
attention}: Inter-episodes attention is where the agent accesses the information from previous episodes, \Cref{fig:attn_score_iea}.
  \item \textbf{Episode-invariant attention}: The agent is able to attend to certain tokens which do not change on changing the episode, \Cref{fig:attn_score_eps_invariant}. 
  \item \textbf{Zero attention}: Some heads have 0 attention scores for all tokens which would not be possible with the vanilla attention.
\end{itemize}

\begin{figure*}[t]
    \centering
    \includegraphics[width=\textwidth]{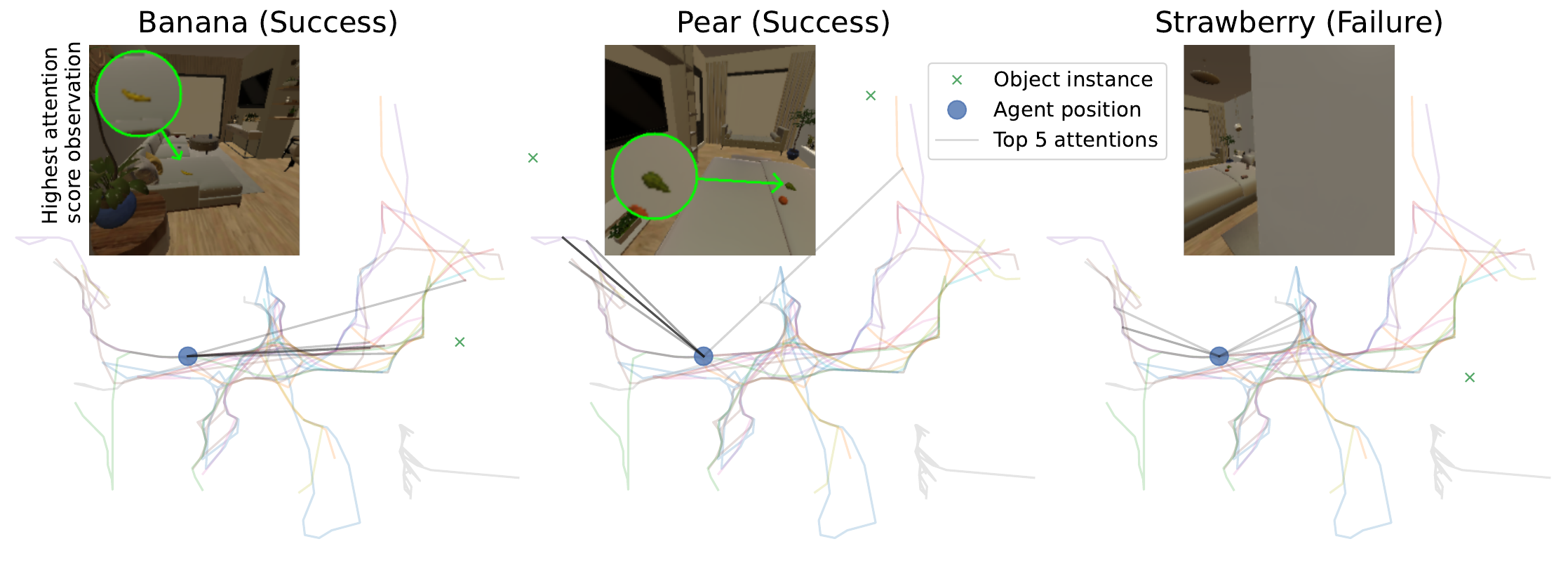}
    \vspace{-15pt}
    \caption{Visualization of an inter-episode attention head, see \Cref{sec:app-what-agent-see}. The colored curves are the trajectories of previous episodes. The blue circle is the agent's position. The green Xs are the instances of the target object type. The black lines represent the agent's attention when the target is the object type mentioned above the image. The lines connect the agent with the point in history that it attends to, the opacity of the line represents the attention score. The overlaid image is visual observation with the highest attention score.}
    \label{fig:trajectory_object_attn_4s}
    \vspace{-15pt}
\end{figure*}

We further analyze the attention pattern between successful and failure episodes. We collect 2455 steps in a trial and then probe the agent's attention scores by querying each object type by adding a new observation with the desired object type at the final step. \Cref{fig:trajectory_object_attn_4s} shows that the agent is able to recall multiple instances of the target object types in its history.

\begin{figure*}[t]
    \centering
    \includegraphics[width=\textwidth,height=0.9\textheight,keepaspectratio]{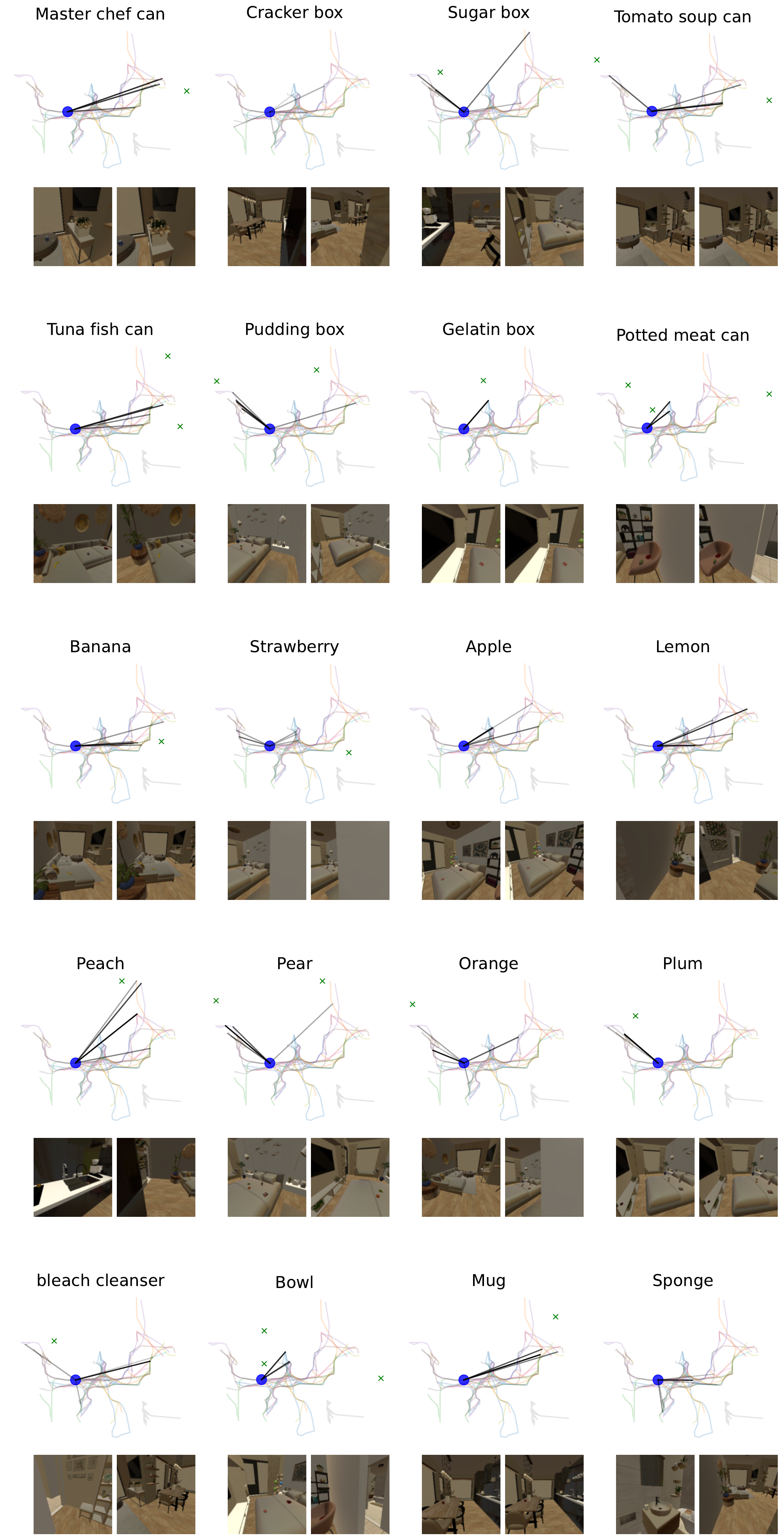}
    \caption{Attention scores of the object detection head described in \Cref{sec:app-what-agent-see}. The colored curves are the trajectories of previous episodes. The blue circle is the agent's position. The black lines represent the agent's attention when the target is the type in above the image. The lines connect the agent with the point in history that it attends to, the opacity of the line represents the attention score. The two images with highest attention score are shown in the 3rd row.}
    \label{fig:app_trajectory_object_attn_4s}
\end{figure*}

\Cref{fig:app_trajectory_object_attn_4s} shows the attention scores for all 20 object types when selected in the 1st step of a new episode after 19 episodes.

\subsection{Impact of Episode Shuffling}
\label{sec:shuffle-abl} 
\begin{figure*}[t]
  \centering
    \includegraphics[width=0.5\textwidth]{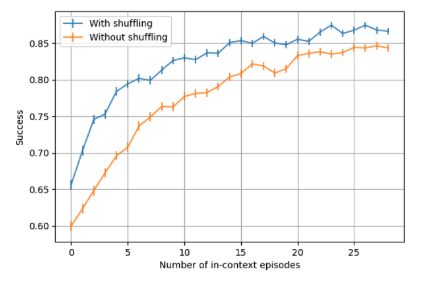}
    \caption{The effect of shuffling in-context episodes during the training}
  \label{fig:shuff}
\end{figure*}

We ran \method on ReplicaCAD with and without in-context episodes shuffling. \Cref{fig:shuff} shows that \method marginally suffers at in-context learning (ICL) when not shuffling episodes in the context during training. Specifically, the final ICL performance has a 3\% lower success rate and the ICL is less efficient. We believe that shuffling the episodes in the context during the training acts as regularization since it creates diverse contexts.

\subsection{Ablations in Miniworld and Darkroom}
\label{sec:miniworld_abls} 
\begin{figure*}
  \centering
  \begin{subfigure}[t]{0.45\columnwidth}
    \includegraphics[width=\textwidth]{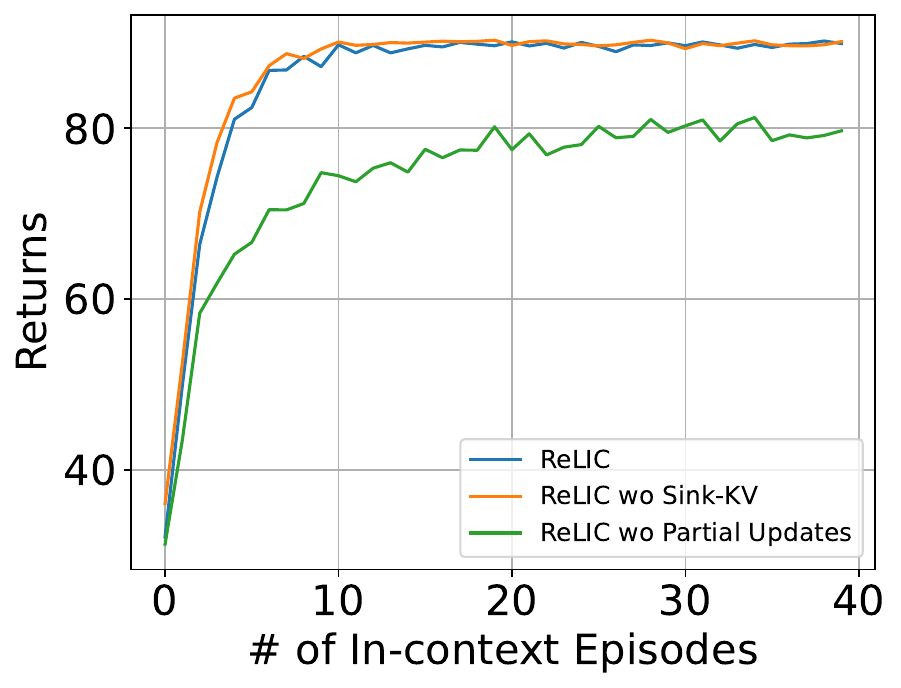}
    \caption{Darkroom}
    \label{fig:dark_ablation}
  \end{subfigure}
  \begin{subfigure}[t]{0.45\columnwidth}
    \includegraphics[width=\textwidth]{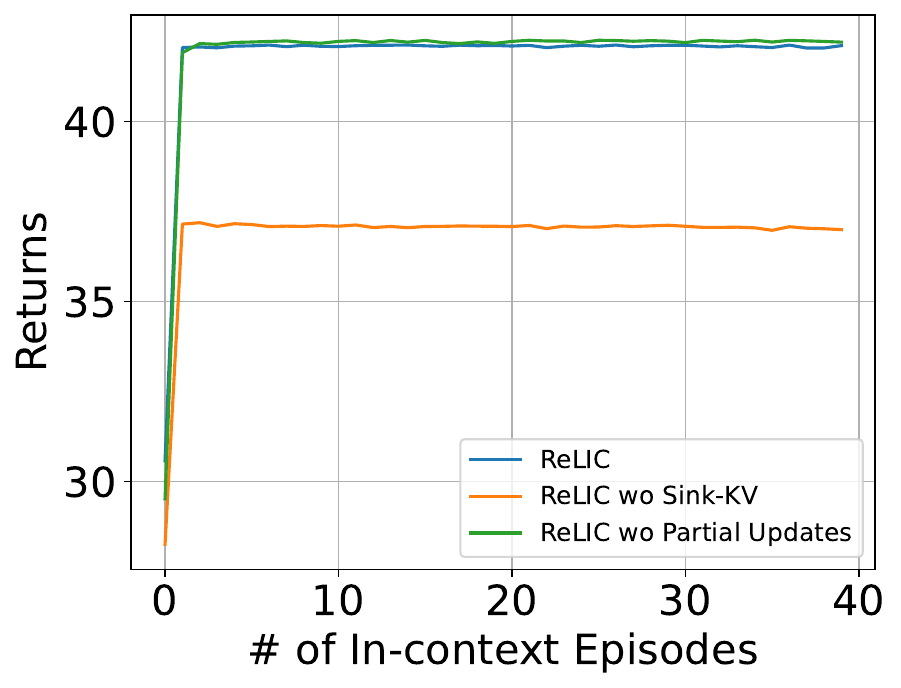}
    \caption{Miniworld}
    \label{fig:mini_ablation}
  \end{subfigure}
  \caption{
    Ablating \method components on Darkroom and Miniworld.} 
  \label{fig:dark_n_mini_ablation}
\end{figure*}

In this section, we run the partial udpate and \sinkv ablations from \Cref{sec:ablation} on the Darkroom and Miniworld tasks from \Cref{sec:darkroom}. 
The result shows that different components in \method is crucial for different tasks while using \method is as good as or better than \method without its components. The ablation shows that Partial Updates is crucial for long horizon tasks like Darkroom and \task as shown in \cref{fig:dark_ablation,fig:updates}, which have horizon of 100 and 500 steps respectively, but not important for short horizon tasks like Miniworld, which is 50 steps, as shown in \cref{fig:mini_ablation}. It also shows that \sinkv is important for tasks with rich observations like Miniworld and \task, which are visual tasks, compared to the Darkroom, which is a grid world task.

\subsection{Training with 64k Context Length}
\label{sec:app-64k-train-eval}

In the main experiment, we showed that we can train on 4k steps and inference for 32k steps. In this experiment, we show that our method \method is able to train with 64k sequence length. We used the same hyperparameters in the main experiment, except the training sequence length which we set to 64k and the number of updates per rollout is increased so that we do updates every 256 steps, same as the main experiment. 
The result in \Cref{fig:app-64k-training} shows that the model is able to learn and generalize on 64k sequence length.

\section{Visual encoder finetuning}
\label{sec:visual-encoder-finetuning}

\begin{figure*}[t]
  \centering
    \includegraphics[width=0.5\textwidth]{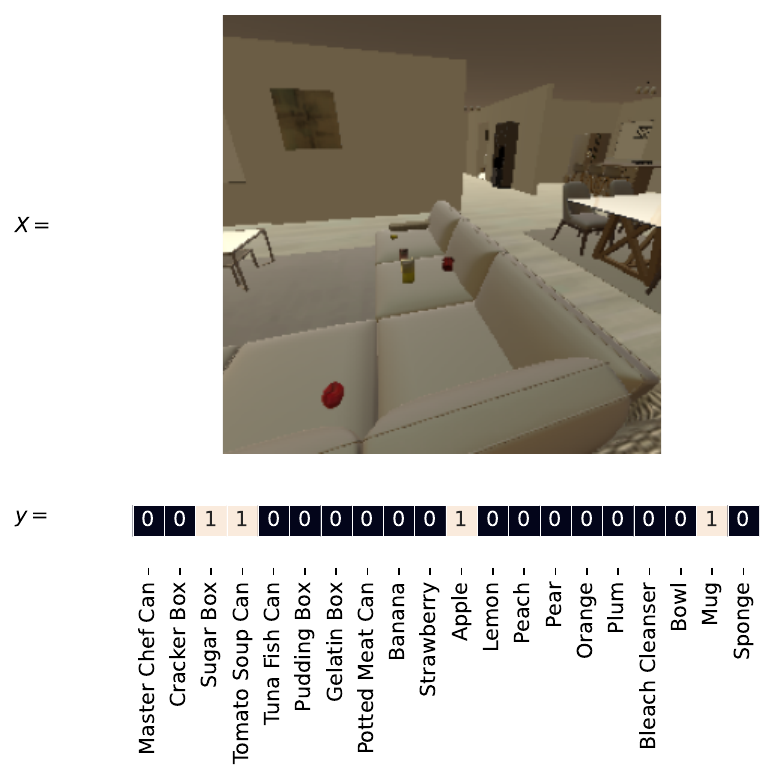}
    \caption{Sample of the finetuning data}
  \label{fig:finetuning_data}
\end{figure*}

We finetuned the visual encoder on a generated supervised task before freezing it to be used in our experiments. 
Each sample, \Cref{fig:finetuning_data}, in the data is generated by placing the agent in front of a random object then the RGB sensor data is used as input $X$. The output $y$ is a binary vector of size 20, the number of available object types, where each element represents whether the corresponding object type is in the image or not. The object type is considered in the image if there is an instance of this object in the image with more than 10 pixels. 
21k samples are generated from the training scenes and object arrangements. The 21k samples are then split to training and validation data with ratios 90\% to 10\%.

The VC-1 model is finetuned using the Dice loss \cite{Sudre_2017_diceloss} by adding a classification head to the output of `[CLS]' token using the generated data. The classification head is first finetuned for 5 epochs with $LR=0.001$ while the remaining of the model is frozen. Then the model is unfrozen and finetuned for 15 epochs with $LR=0.00002$.
\gc{Maybe add a citation for this strategy here. \\ Ahmad: What strategy? }

\section{Sink KV}
\label{sec:app-sink-kv}

\begin{figure*}[b]
  \centering
  \begin{subfigure}[t]{0.45\columnwidth}
    \includegraphics[width=\textwidth]{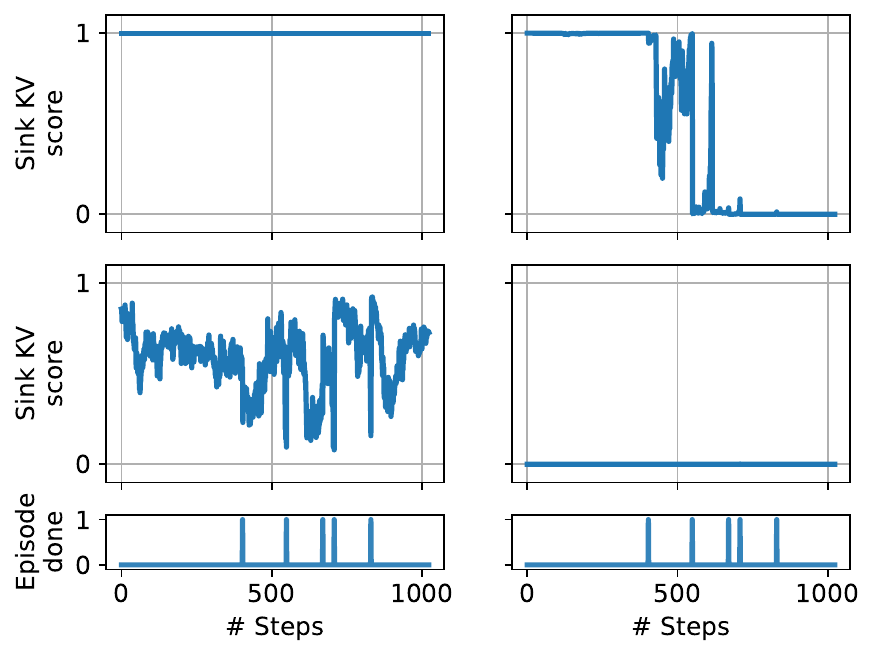}
    \caption{Different patterns of Sink $KV$ scores  for 1k input tokens.}
    \label{fig:app-sink-kv-scores}
  \end{subfigure}
  \begin{subfigure}[t]{0.45\columnwidth}
    \includegraphics[width=\textwidth]{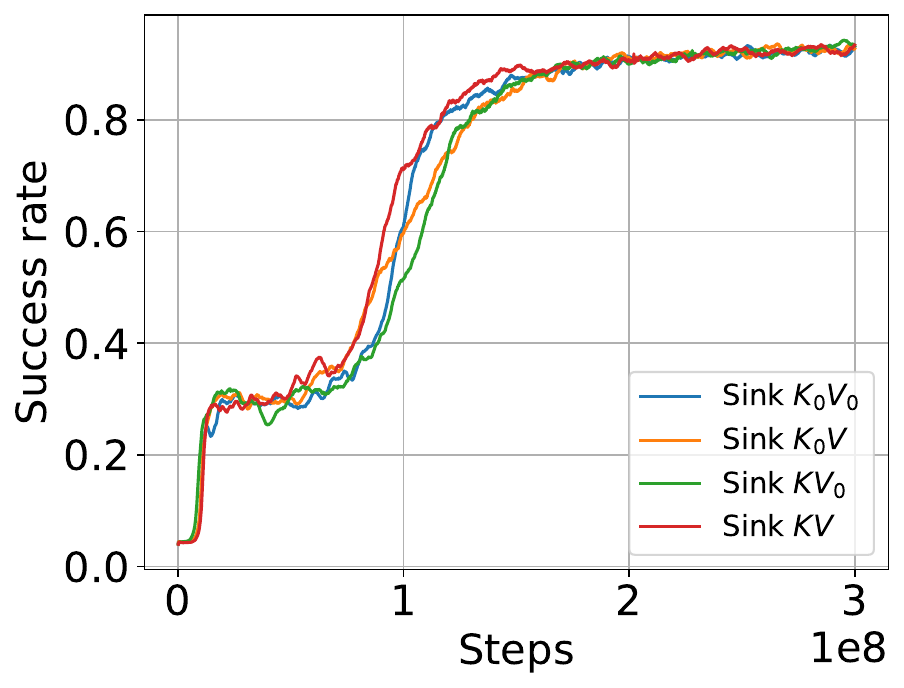}
    \caption{The learning curve for the Sink $KV$ variants.}
    \label{fig:app-sink-kv-variants}
  \end{subfigure}
  \caption{
    Sink $KV$ analysis.
  }
  \label{fig:sink-kv-analysis}
\end{figure*}

We introduce Sink $KV$, a modification to the attention calculation in the attention layers.
We first describe the vanilla attention \cite{vaswani2023attention}, the issue and the motivation to find a solution. Then we discuss the proposed solutions and introduce the \sinkv technique. Finally, we anlayze different variants of \sinkv.

\subsection{Motivation}
The vanilla attention is the component responsible for the interaction between the tokens in the sequence. The output for each token is calculated by weighting the value of all tokens.
The input to the attention layer is the embeddings of the input tokens $E\in \mathbb{R}^{n\times d}$ where $n$ is the number of input tokens and $d$ is the dimension of the embeddings.

First the embeddings $E$ are linearly projected to the Key $K$, Value $V$ and Query $Q$. Then the attention scores are calculated using $S=\text{Softmax}(QK^T/\sqrt{d_k})$ where $d_k$ is the dimensions of the keys. The output $A$ is calculated as a weighted sum of the values $V$,  $A=SV$. 

The calculation of the attentions scores $S$ using the $\text{Softmax}$ forces the tokens to attend to values $V$, even if all available values do not hold any useful information, since the sum of the scores is 1 \citep{off_by_one_softmax}. This is especially harmful in cases where the task requires exploration. As the agent explores more, a more useful information may appear in the sequence. If the agent is forced to attend to low information tokens at the beginning of the exploration, it will introduce noise to the attention layers.

\subsection{Solutions}
\label{sec:app-sinkkv-solutions}

Softmax One from \cite{off_by_one_softmax} addresses this issue by adding $1$ to the denominator of the $\text{Softmax}$, $\text{Softmax}_1(x_i) :=\exp(x_i)/(1+\sum_j{\exp(x_j)})$, which is equivalent to having a token with $k=0$ and $v=0$. This gives the model the ability to have 0 attention score to all tokens, we refer to Softmax One as Sink $K_0V_0$.

Sink tokens from \cite{xiao2023efficient} are another approach to address the same issue by prepending learnable tokens to the input tokens $E=[E_s \circ E_{input}]$ where $E$ is the input embedding to the model and $[A\circ B]$ indicates concatenation along the sequence dimension of the $A$ and $B$ matrices .

\sinkv is a generalization of both approaches. It modifies the attention layer by adding a learnable Key $K_s\in\mathbb{R}^{n\times d_k}$ and values $V_s\in\mathbb{R}^{n\times d}$. In each attention layer, we simply prepend the learnable $K_s$ and $V_s$ to the vanilla keys $K_v$ and values $V_v$ to get the $K=[K_s \circ K_v]$ and $V=[V_s \circ V_v]$ used to calculate the attention scores then the attention output. 

In the case $K_s=0$ and $V_s=0$, \sinkv becomes equivalent to Softmax One. It can also learn the same $K$s and $V$s corresponding to the Sink Token since our model is casual and the processing of the Sink Token is not affected by the remaining sequence.

\subsection{\sinkv variants}
We tried a variant of \sinkv where the either the Value or the Key is set to 0, referred to as Sink $KV_0$ and Sink $K_0V$ respectively. All variants perform similarly in terms of the success rate as shown in \Cref{fig:app-sink-kv-variants}.

\Cref{fig:app-sink-kv-scores} shows different patterns the model uses the Sink $KV_0$. The model can assign all attention scores to the Sink $KV_0$, which yields a zero output for the attention head, or assign variable scores at different time in the generation. For example, one the attention heads is turned off during the 1st episode of the trial by assigning all attention score to the Sink $KV_0$ then eventually move the attention to the input tokens in the new episodes. The model is also able to ignore the Sink $KV_0$ by assigning it 0 attention scores as shown in the figure. 

\end{document}